\documentclass{article}

\usepackage{amsmath}
\usepackage{amssymb}
\usepackage{amsfonts}
\usepackage{mathtools}
\usepackage{placeins}
\usepackage{rotating}
\usepackage{bm}
\usepackage{tikz}
\usetikzlibrary{
    positioning,
    arrows.meta,
    shapes.geometric,
    fit,
    backgrounds,
    calc
}
\usepackage[left=1.25in, right=1.25in, top=1in, bottom=1in]{geometry}
\usepackage{graphicx}
\usepackage{subcaption}
\usepackage{booktabs}
\usepackage{multirow}
\usepackage{array}
\usepackage{dsfont}
\usepackage{makecell}
\usepackage{siunitx}
\usepackage{hyperref}
\usepackage{url}
\usepackage{xspace}
\usepackage{tikz}
\usepackage{natbib}
\usetikzlibrary{positioning,arrows.meta,shapes.geometric,fit,backgrounds}

\title{Latent Flow Matching for Molecular Graph Generation}
\author{
\textbf{Mathis Goupillon}\footnote{equal contribution}\\
École normale supérieure Paris-Saclay
\and
\textbf{Roman Bresson}$^*$ \\
Mohamed bin Zayed University of Artificial Intelligence
\and
\textbf{Konstantinos Divriotis}\\
Mohamed bin Zayed University of Artificial Intelligence
\and
\textbf{Michalis Vazirgiannis}\\
Mohamed bin Zayed University of Artificial Intelligence\\
LIX, CNRS, École Polytechnique, IP Paris
}

\date{}

\begin{document}

\maketitle

\begin{abstract}
Modern graph generative models typically operate directly in the discrete graph space, explicitly generating node and edge variables, which can become costly as graphs grow. In this paper, we perform generation explicitly on latent representations of entire graphs obtained from a pretrained Variational Autoencoder with high reconstruction fidelity. The generated representations, obtained through flow matching, are then decoded only at the final step. Across molecular benchmarks of increasing size, our approach achieves strong validity and FCD while offering a favorable quality-efficiency trade-off compared with state-of-the-art explicit graph generative models. One of the main advantages of this formulation is that the graph representation only needs to be learned once, after which the same one can be reused across multiple generative objectives without retraining. We demonstrate generation guided by molecular properties and further introduce validity-aware generation though a classifier learned directly in latent space. All code will be made available upon acceptance.
\end{abstract}
\section{Introduction}
\label{sec:introduction}

Recent advances in graph generation have been largely driven by diffusion and flow-based models operating directly on explicit node and edge representations \citep{vignac2023digress,qinmadeira2024defog,roos2026categorical}. These methods model the graph distribution in its original discrete space and have achieved strong results across several graph generation benchmarks, but they suffer from having to manipulate large objects (full adjacency matrices and node embeddings) as well as being potentially permutation-sensitive. An alternative line of work instead learns compact, permutation-invariant graph-level representations via autoencoder architectures \citep{pigvae,grale,bresson2026graviti,autoencoders_basic}, decoupling representation learning from distribution modeling (see Section~\ref{sec:background} for details). One main drawback is that such architectures are known to yield poor generative performance without domain-specific knowledge being injected \citep{juncvae,gen3}. 

This raises the question of whether an approach could leverage the best of both world, by moving distribution modeling entirely into such a continuous graph-level representation space; doing so should offer a competitive alternative to explicit graph generation in terms of sample quality, distributional fidelity, while retaining computational efficiency. This question is the central problem adressed in this work. Beyond unconditional generation, we ask whether the same latent representation can support controlled generation toward prescribed graph-level properties, and whether generation can be further biased toward latent regions that decode more reliably, improving chemical validity. 

Our contributions are threefold. First, we train a flow-matching~\citep{lipman2022flow} model to learn the latent distribution of the of a pretrained high-fidelity graph-level autoencoder. Second, we combine conditional flow matching with gradient-based guidance from differentiable property predictors, and demonstrate accurate control of logP and molecular weight while maintaining high validity and diversity on molecular datasets. Finally, we introduce an endpoint-based validity-aware flow matching formulation, based on Variational flow  ~\citep{eijkelboom2024vfm}, that substantially improves chemical validity and characterize the resulting trade-off between validity and distributional fidelity.

\section{Background and Related Work}
\label{sec:background}

\paragraph{Flow matching and generative modeling}
\label{sec:vfm-background}

Generative modeling is the process of learning how to sample from an unknown data distribution $q$, so as to obtain novel, valid and varied data points. A usual process is to sample a point $z$ from a known and simple prior $p$ and learn a transformation $T$ so that $T(z)\sim q$. In a VAE \citep{kipf_variational_2016}, $T$ is simply the decoder. More advanced methods, such as diffusion \citep{dhariwal2021diffusion} and flow matching \citep{lipman2022flow} rely on applying a multi-step process, where each step removes noise and injects signal. In particular, flow matching learns this distribution by regressing a time-dependent vector field $v_\theta(z_t, t)$ that transports samples from a simple source distribution (e.g., a standard Gaussian) to $p(z)$ along a prescribed probability path, avoiding the iterative denoising schedules of diffusion models while retaining comparable sample quality. Conditional generation \citep{li2025towards} extends this formulation by conditioning $q$ on desired properties of the generated object (i.e. a prompt). Flow matching enables this by conditioning the learned vector field directly \cite{lipman2024flowmatchingguidecode}.

\paragraph{Graph-Level Continuous Representations}

Let $\mathcal{G}$ denote the graph space, where a graph with $n$ nodes is represented as an $n\times c_n$ matrix of node features and an adjacency tensor of shape $n\times n \times c_e$ representing edge classes. In contrast to node-level autoencoders \cite{kipf_variational_2016} (which encodes the graph as a permutation-equivariant $n\times d$ matrix), a graph-level encoder $E$ maps a complete graph $G \in \mathcal{G}$ to a fixed-dimensional permutation-invariant representation $z = E(G) \in \mathbb{R}^d$, while a decoder $D$ maps this representation back to graph space through $\hat{G} = D(z)$, with $\hat{G}$ an attempt at reconstructing $G$. Several recent architectures pursue this direction such as PIGVAE \citep{pigvae}, GRALE \citep{grale}, or GraViti \citep{bresson2026graviti} differing primarily in how permutation invariance is enforced and in the fidelity of reconstruction on larger, more heterogeneous graphs. These approaches, and more particularly GraViti, demonstrate that compact continuous representations can retain sufficient structural information for near-exact graph reconstruction on molecular datasets of different scales. In particular, PigVAE and GraViti provide a variational autoencoder architecture~\citep{autoencodingvariational}, which structures the latent space explicitly with downstream generation in mind (Section~\ref{sec:method}); nonetheless, while cheap, one-shot VAE sampling is known to fail at generating consistently valid molecules \citep{Grammar,juncvae,Hu2023-xz,NGUYEN20253867}, making it a poor generative model on its own.

\paragraph{Generation in explicit graph space.}
Most recent graph generative models operate directly on node and edge variables, keeping the graph representation explicit throughout generation. DiGress~\citep{vignac2023digress} applies discrete denoising diffusion to categorical graph representations, while CatFlow~\citep{eijkelboom2024vfm} and DeFoG~\citep{qinmadeira2024defog} similarly formulate generation directly in graph space through categorical or discrete Flow Matching. More recently, Categorical Flow Maps~\citep{roos2026categorical} introduced continuous trajectories over categorical variables by evolving them in the probability simplex, enabling few-step generation. Despite this continuous formulation, the generative state still explicitly represents node and edge categories rather than a learned representation of the graph itself.

\paragraph{Generation through continuous representations.}

While latent generation approaches for graphs exist, they usually operate on node-level representations \citep{10508504,zhou2024unifying,siraudin2026principled,xu2023geometric} due to the only recent emergence of high-fidelity graph-level approaches. These methods include highly specialized VAEs leveraging domain-dependant constraints~\citep{gomez2018automatic,juncvae} offering high validity, but with capped distributional fidelity \citep{vignac2023digress}. Building on those graph-level latents, ChemFlow~\citep{wei2024chemflow} has shown interesting latent navigation, without being a generative model, while NGG~\citep{evdaimon2024neural} employs a small variational graph autoencoder on general graphs based on structure conditioning, limiting its applicability to more complex data.

\begin{figure*}[t]
\centering

\begin{tikzpicture}[
    scale=0.72,
    transform shape,
    font=\small,
    >=Latex,
    data/.style={draw, rounded corners=2pt, minimum width=2.2cm, minimum height=0.9cm, align=center, fill=green!10},
    latent/.style={draw, rounded corners=2pt, minimum width=2.4cm, minimum height=0.9cm, align=center, fill=violet!10},
    model/.style={draw, rounded corners=2pt, minimum width=2.4cm, minimum height=1.0cm, align=center, fill=blue!10},
    process/.style={draw, rounded corners=2pt, minimum width=2.3cm, minimum height=0.9cm, align=center, fill=orange!10},
    dataset/.style={draw, rounded corners=2pt, minimum width=2.cm, minimum height=0.95cm, align=center, fill=gray!10},
    decision/.style={draw, diamond, aspect=2, align=center, inner sep=1.5pt, fill=yellow!12},
    encoder/.style={trapezium, trapezium left angle=70, trapezium right angle=70, shape border rotate=270, draw, minimum width=1.8cm, minimum height=1.3cm, align=center, fill=red!10},
    decoder/.style={trapezium, trapezium left angle=70, trapezium right angle=70, shape border rotate=90, draw, minimum width=1.8cm, minimum height=1.3cm, align=center, fill=red!10},
    arrow/.style={-{Latex[length=2mm]}, thick},
    guidancearrow/.style={-{Latex[length=2mm]}, thick, dashed, red!70!black},
    paneltitle/.style={font=\large\bfseries, fill=white, inner xsep=6pt, inner ysep=2pt}
]

\node[data] (inputgraph) at (0,0) {Input graph\\$G$};
\node[encoder] (encoder) at (3,0) {GraViti\\encoder};
\node[latent] (latent) at (6,0) {Graph-level\\latent space};
\node[decoder] (decoder) at (9,0) {GraViti\\decoder};
\node[data] (outputgraph) at (12,0) {Generated\\graph $\hat{G}$};

\draw[arrow] (inputgraph) -- (encoder);
\draw[arrow] (encoder) -- (latent);
\draw[arrow] (latent) -- (decoder);
\draw[arrow] (decoder) -- (outputgraph);

\node[process] (source) at (6,3.5) {Source\\$z_0 \sim \mathcal{N}(0,I)$};
\node[model] (fm) at (6,2) {\textbf{Latent FM}\\$v_\theta(z_t,t)$};

\draw[arrow] (source) -- (fm);
\draw[arrow] (fm) -- (latent);

\node[process] (guidance) at (9,2) {Validity\\guidance};

\draw[guidancearrow] (guidance.west) -- (fm.east);

\begin{scope}[on background layer]
\node[
    draw=orange!70!black,
    rounded corners=6pt,
    inner sep=10pt,
    fit=(source)(fm)(guidance)(latent)(decoder)(outputgraph)
] (topbox) {};
\end{scope}

\node[paneltitle, text=orange!70!black] at (8.8,4.6) {Latent generation};


\node[model] (predictor) at (16.2,2)
{\textbf{Validity predictor}\\$s_\omega(z_t,t)$};

\node[dataset] (balanced) at (16.2,0.4)
{Balanced validity dataset\\$50\%$ valid / $50\%$ invalid};

\draw[arrow] (balanced.north) -- node[right, font=\scriptsize] {train} (predictor.south);

\draw[guidancearrow] (predictor.west) -- (guidance.east)
node[midway, above, font=\scriptsize, align=center] {predictor used\\at sampling};


\node[process] (source2) at (0,-3.5) {Source\\$z_0 \sim \mathcal{N}(0,I)$};
\node[model] (fm2) at (3,-3.5) {\textbf{Latent FM}\\$v_\theta(z_t,t)$};
\node[latent] (zgenerated) at (6,-3.5) {Generated\\latent $z$};
\node[decoder] (decoder2) at (9,-3.5) {GraViti\\decoder};
\node[data] (molecule) at (12.2,-3.5) {Decoded\\molecule};

\draw[arrow] (source2) -- (fm2);
\draw[arrow] (fm2) -- (zgenerated);
\draw[arrow] (zgenerated) -- (decoder2);
\draw[arrow] (decoder2) -- (molecule);


\node[decision] (validitycheck) at (16.2,-3.5) {Valid?};

\draw[arrow] (molecule.east) -- (validitycheck.west);

\node[dataset] (valid)   at (14.9,-2.0) {Valid\\$v=1$};
\node[dataset] (invalid) at (17.5,-2.0) {Invalid\\$v=0$};

\draw[arrow]
(validitycheck.north) -- ++(0,0.15) -- ++(-1.3,0) -- (valid.south);
\node[font=\scriptsize] at (14.5,-2.8) {yes};

\draw[arrow]
(validitycheck.north) -- ++(0,0.15) -- ++(1.3,0) -- (invalid.south);
\node[font=\scriptsize] at (17.75,-2.8) {no};


\draw[thick] (valid.north)   -- (14.9,-1.25) -- (16.2,-1.25);
\draw[thick] (invalid.north) -- (17.5,-1.25) -- (16.2,-1.25);
\fill (16.2,-1.25) circle (1.5pt);
\draw[arrow] (16.2,-1.25) -- (balanced.south);


\begin{scope}[on background layer]
\node[
    draw=gray!65,
    rounded corners=6pt,
    inner sep=10pt,
    fit=(source2)(fm2)(zgenerated)(decoder2)(molecule)
        (validitycheck)(valid)(invalid)
] (bottombox) {};
\end{scope}

\node[paneltitle] at (8.6,-1.35) {Validity predictor construction};


\draw[dashed, gray!65] (decoder.south) -- ++(0,-0.9) -- (decoder2.north);
\node[gray!65, font=\scriptsize] at (9.9,-2.3) {shared decoder};

\end{tikzpicture}

\caption{
Overview of latent graph generation and validity-guided sampling. Top: latent flow matching; Bottom: construction of the validity dataset and validity predictor training.
}
\label{fig:pipeline}

\end{figure*}
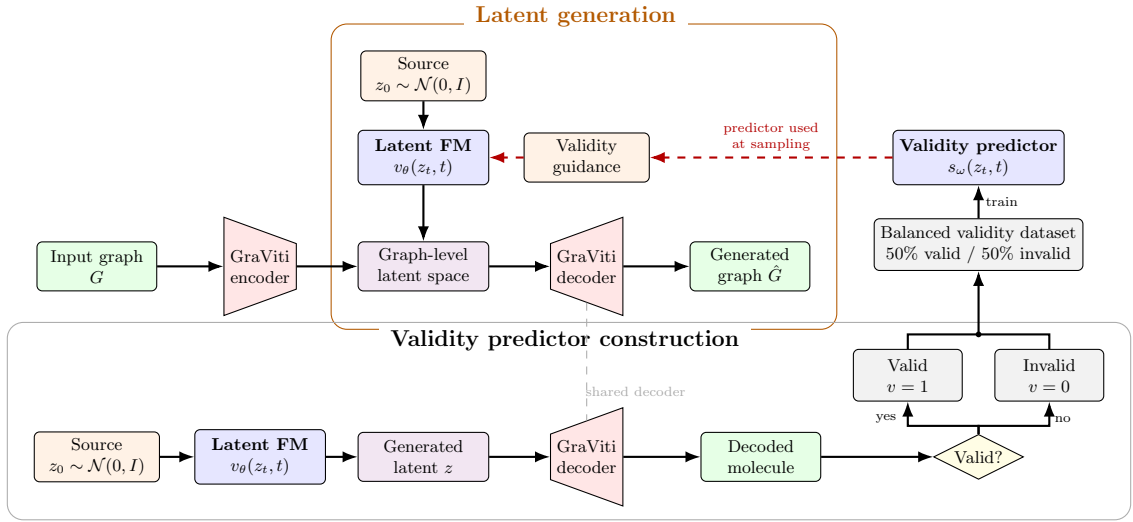
\section{Method}
\label{sec:method}

\paragraph{Latent Flow Matching}

We formulate graph generation as distribution learning in a graph-level continuous representation space. Representations are obtained from a pretrained GraViti-based variational autoencoder~\citep{bresson2026graviti}, since it is freely available and trainable and achieves high reconstruction quality.  The encoder and decoder are considered as pretrained, and remain fixed, decoupling graph representation learning from generative modeling.

Let $E:\mathcal{G}\to\mathbb{R}^d$ denote the graph encoder of our pretrained VAE and $z=E(G)\in\mathbb{R}^d$ the latent representation of a graph $G$. We learn the latent graph distribution using Flow Matching~\citep{lipman2022flow}, transporting an isotropic Gaussian source distribution $z_0\sim\mathcal{N}(0,I)\in\mathbb{R}^d$ toward latent graph embeddings $z_1\sim p_{\mathrm{data}}\in\mathbb{R}^d$, where $p_{\mathrm{data}}$ is the empirical distribution of encoder outputs obtained by applying $E$ to the training graphs. We adopt the standard Flow Matching formulation, detailed in Appendix~\ref{app:flowmatching}. To improve the coupling between source and target samples, we employ mini-batch optimal transport~\citep{tong2024improving}, which promotes shorter and straighter transport trajectories while preserving the source and target marginals.

In order to avoid data leakage from the VAE's training, we adopt the same training, validation and test splits that are provided with GraViti. Therefore, both our model and the pretrained VAEs are trained on the same molecules exactly. We then learn a neural flow model over the distribution of graph representations, such that the complete generative trajectory is modeled in continuous latent space and only the final representation is decoded back to graph space. At generation time, the learned field is integrated from $t=0$ to $t=1$ and the resulting latent endpoint is decoded into a graph only after the last step.

\paragraph{Validity Guidance} We further exploit the differentiability of the latent dynamics to promote molecule validity. After an unconditional neural flow model is trained, a time-dependent classifier $s_{\omega}$ is trained on valid and invalid latent states. The classifier is time-dependent because, given an intermediate latent state \(z_t\) and its flow time \(t\), it predicts whether the corresponding sample is expected to decode into a valid graph at the end of the trajectory. To train this prediction throughout the flow, intermediate states are constructed along linear trajectories between noise samples and valid or invalid endpoint latents, with each state inheriting the validity label of its endpoint. Positive examples are obtained from encoded valid training graphs, while negative examples are generated by the latent Flow Matching model and retained when they decode into invalid graphs. The construction of this dataset is critical to obtaining informative guidance: rather than relying on easily separable negatives far from the graph latent distribution, we use hard negatives lying in regions actually explored by the generator. This encourages the classifier to learn the validity boundary relevant to the sampling trajectories. Following the principle of classifier guidance~\citep{dhariwal2021diffusion},
sampling can be steered according to
\begin{equation}
    \tilde v_{\theta}(z_t,t)
    =
    v_{\theta}(z_t,t)
    +
    \mathds{1}[t \geq t_{\mathrm{start}}]\lambda_{\mathrm{val}}
    \nabla_{z_t}
    \log p_{\omega}(\mathrm{valid}\mid z_t,t).
\end{equation}
Where $p_w$ is the validity probability obtained from $s_w$ and $\lambda_{\mathrm{val}}$ is a parameter that controls the steering intensity. For stability and diversity, we restrict guidance to the later part of the trajectory (after $t_\text{start}$), where samples approach the learned graph distribution.

\paragraph{Property-Controlled Generation}

A desirable property of generative models is the ability to be prompted by properties of the generated molecule. We condition the flow matching field on a target property vector $c$, yielding $v_\theta(z_t,t,c)$, and complement this conditioning with a differentiable property predictor $r_\omega(z_t,t)$ trained on intermediate states of the generative trajectories (using a similar process to that of the validity classifier). For a target $c$, the predictor provides an adaptive correction during sampling,
\begin{equation}
    \tilde v_{\theta}(z_t,t,c)
    =
    v_{\theta}(z_t,t,c)
    -
     \mathds{1}[t \geq t_{\mathrm{start}}]\lambda_{\mathrm{prop}} \,
    \nabla_{z_t}
    \mathcal{L}_{\mathrm{prop}}
    \big(r_{\omega}(z_t,t),c\big),
\end{equation}
where $\mathcal{L}_{\mathrm{prop}}$ is the squared difference between the predicted property and the target value $c$, and where guidance is applied only once $t$ exceeds a threshold $t_{\mathrm{start}}$, since $r_\omega$'s gradient is unreliable while $z_t$ remains close to pure noise. Conditioning therefore shapes the global transport toward the requested property, while gradient guidance refines individual trajectories once sufficient structure has emerged. Property control remains entirely within the graph-level latent space and requires no modification of the graph decoder.

\paragraph{Validity-Aware Endpoint Generation}
\label{sec:validity_endpoint}

Validity guidance modifies a pretrained generative field at sampling time. We
also investigate whether validity can be incorporated directly into the
generative distribution. Building on the endpoint formulation of Variational
Flow Matching~\citep{eijkelboom2024vfm}, we model a conditional distribution
over final graph representations,
\begin{equation}
    q_{\theta}(z_1 \mid z_t,t)
    =
    \mathcal{N}\!\left(
        \mu_{\theta}(z_t,t),
        \operatorname{diag}\!\left(\sigma_{\theta}^2(z_t,t)\right)
    \right).
\end{equation}

Let $a_{\omega}(z)=\sigma\!\left(s_{\omega}(z,1)\right)$ denote the validity score produced by a frozen classifier. When interpreted as
an estimate of $P(V=1\mid z)$, where $V$ denotes graph validity, the joint
density of an endpoint and the validity event factorize as
\begin{equation}
    p_{\theta,\omega}(z_1,V=1\mid z_t,t)
    =
    q_{\theta}(z_1\mid z_t,t)\,a_{\omega}(z_1).
\end{equation}
Importantly, $a_{\omega}$ is a conditional validity probability and not a
density over latent space. The probability that an endpoint sampled from the
model is valid is therefore
\begin{equation}
    P_{\theta,\omega}(V=1\mid z_t,t)
    =
    \mathbb{E}_{z_1\sim q_{\theta}}
    \left[a_{\omega}(z_1)\right]
\end{equation}

Motivated by this factorization, we train the endpoint distribution with $\mathcal{L}_{\mathrm{end}}=\mathcal{L}_{\mathrm{NLL}}+\lambda_{\mathrm{val}}\mathcal{L}_{\mathrm{val}},
$ where $\mathcal{L}_{\mathrm{NLL}} =
    \mathbb{E}_{z_1\sim p_{\mathrm{data}}}
    \left[
        -\log q_{\theta}(z_1\mid z_t,t)
    \right]$ and $\mathcal{L}_{\mathrm{val}} = 
    \mathbb{E}_{\tilde z_1\sim q_{\theta}}
    \left[
        -\log a_{\omega}(\tilde z_1)
    \right]$.
The first term preserves likelihood on observed graph representations, while
the second explicitly moves probability mass toward endpoints predicted to
decode into valid graphs. Since $-\log$ is convex, Jensen's inequality gives
\begin{equation}
    \mathcal{L}_{\mathrm{val}}
    \geq
    -\log P_{\theta,\omega}(V=1\mid z_t,t),
\end{equation}
providing a direct probabilistic interpretation of the validity objective. The
samples $\tilde z_1$ are reparameterized through $q_{\theta}$, allowing
validity gradients to update the endpoint distribution.

At inference, the endpoint distribution can be exploited in two ways. Its
predicted mean provides a deterministic endpoint estimate, while stochastic
sampling allows several plausible endpoints to be considered at each step.
We draw $K$ candidates from $q_{\theta}(z_1\mid z_t,t)$ and favor those with
higher predicted validity,
\begin{equation}
    \hat z_1
    =
    \sum_{k=1}^{K} w_k z_1^{(k)},
    \qquad
    w_k
    \propto
    \exp\!\left(
        \beta s_{\omega}(z_1^{(k)},1)
    \right).
\end{equation}
This reweighting exploits the uncertainty of the learned endpoint distribution
while favoring trajectories toward latent regions with higher predicted
validity.
\section{Experiments}
\label{sec:experiments}

\subsection{Experimental Setup}
\label{sec:experimental_setup}

\begin{table}[t]
    \centering
    \caption{Statistics of the used datasets. Numbers of atoms and bonds are reported on average. Last column is the dimension of the latent representation.}
    \resizebox{\linewidth}{!}{
    \begin{tabular}{ c c c c c c c}
        \toprule
         Name & \# Graph & \#  atom & \#  bonds & Max Size & Atom types & Emb. Dim. \\
         \midrule
         QM9 & $130$k & $8.79$ & $9.4$ & $9$ & $4$ & $64$\\
         PubChem16 & $12.89$M & $13.87$ & $14.19$ & $16$ & $31$ & $256$\\
         PubChem32 & $74.34$M & $21.75$ & $21.15$ & $32$ & $31$ & $512$\\
         \bottomrule
    \end{tabular}}

    \label{tab:datasets}
\end{table}

\paragraph{Datasets.}
We evaluate our approach on QM9~\citep{ramakrishnan2014qm9} and PubChem \citep{PubChem}, two molecular graph datasets with substantially different scales and structural complexity. For PubChem, we use two subsets clipped at a maximum size of $16$ and $32$ heavy atoms respectively. All datasets have $4$ edge types, corresponding to simple, double, triple bonds, and one extra class for "no bond". Statistics are described in Table \ref{tab:datasets}.

\paragraph{Baselines.}
We compare latent generation against explicit graph generative models: the diffusion-based DiGress~\citep{vignac2023digress}, a the flow-based Categorical Flow Maps~\citep{roos2026categorical}. We additionally report Gaussian sampling from the VAE prior as a latent-space baseline, allowing us to quantify the benefit of learning the latent graph distribution with flow matching. All models are trained and evaluated on the same splits.

\paragraph{Evaluation protocol.}
For each configuration, we perform $20$ independent generation runs of $1{,}024$
molecules, for a total of $20{,}480$ generated samples. 
Implementation details and hyperparameters are provided in appendix \ref{app:training_settings}.

\subsection{Unconditional Graph Generation and Validity Guidance}
\label{sec:unconditional_generation}

\begin{table}[t]
\centering
\caption{
Unconditional generation results on our three benchmarks. Validity, novelty and uniqueness are reported as mean $\pm$ standard deviation over 20 runs of 1,024 molecules. FCD is computed over the generated set. Time denotes end-to-end generation of 1,024 molecules and includes the VAE decoding for our method. NFE stands for number of function evaluations (1 per step for Euler, 2 per step for midpoint and Heun solver). The numerical solvers are detailed in Appendix~\ref{app:ode_solvers}. Our approaches are Latent FM, Latent FM with validity guidance (VG) and Validity-Aware VFM (VA-VFM). Qualitative examples of pc32 generation are provided in Appendix~\ref{app:pc32_qualitative}.}
\label{tab:uncond}
\resizebox{\linewidth}{!}{
\begin{tabular}{llccccccc}
\toprule
\textbf{Method} &
\textbf{Sampler} &
\textbf{NFE} &
\textbf{Validity } $\uparrow$ &
\textbf{Uniq. } $\uparrow$ &
\textbf{Novelty } $\uparrow$ &
\textbf{FCD} $\downarrow$ & 
\textbf{Time (s)} $\downarrow$ &
\textbf{VUN/T} $\uparrow$ \\
\midrule
\textbf{QM9}\\
\midrule
DiGress
& Diffusion & 2
& $47.3 \pm 1.4$
& $97.5 \pm 0.9$
& N/A
& $3.420$
& $0.324 \pm 0.081$ 
& $1457.5$\\

DiGress
& Diffusion & 5
& $76.7 \pm 1.7$
& $99.5 \pm 0.3$
& N/A
& $1.150$
& $0.633 \pm 0.005$ 
& $1234.6$\\

DiGress
& Diffusion & 20
& $97.8 \pm 0.4$
& $\textbf{99.7} \pm 0.2$
& N/A
& $0.087$
& $2.275 \pm 0.030$ 
& $438.9$ \\

DiGress
& Diffusion & 500
& $\textbf{99.9} \pm 0.1$
& $99.6 \pm 0.2$
& N/A
& $\textbf{0.061}$
& $54.801 \pm 0.13$
& $18.6$ \\

\midrule

CFM-CSD
& CSD & 1
& $94.6 \pm 0.6$
& $98.6 \pm 0.6$
& N/A
& $1.873$
& $0.391 \pm 0.005$ 
& $2442.8$
\\

CFM-CSD
& CSD & 2
& $96.5 \pm 0.4$
& $\textbf{99.7} \pm 0.1$
& N/A
& $0.280$
& $0.446 \pm 0.002$ 
& $2209.0$
\\

CFM-CSD
& CSD & 5
& $98.8 \pm 0.3$
& $\textbf{99.7} \pm 0.2$
& N/A
& $0.111$
& $0.632 \pm 0.002$ 
& $1596.0$
\\

CFM-CSD
& CSD & 20
& $99.6 \pm 0.2$
& $99.6 \pm 0.2$
& N/A
& $0.082$
& $1.520 \pm 0.002$ 
& $668.3$ \\

CFM-CSD
& CSD & 500
& $99.7 \pm 0.1$
& $99.5 \pm 0.1$
& N/A
& $0.069$
& $30.129 \pm 0.194$ 
& $33.71$ \\

\midrule

CFM-ECLD
& ECLD & 1
& $97.7 \pm 0.4$
& $97.3 \pm 0.5$
& N/A
& $1.791$
& $0.401 \pm 0.003$ & $2427.5$ \\

CFM-ECLD
& ECLD & 2
& $98.2 \pm 0.4$
& $99.6 \pm 0.2$
& N/A
& $0.480$
& $0.447 \pm 0.002$ & $2240.6$ \\

CFM-ECLD
& ECLD & 5
& $99.2 \pm 0.3$
& $99.6 \pm 0.2$
& N/A
& $0.327$
& $0.626 \pm 0.003$ & $1616.2$ \\

CFM-ECLD
& ECLD & 20
& $99.5 \pm 0.3$
& $99.6 \pm 0.2$
& N/A
& $0.342$
& $1.522 \pm 0.002$ & $666.8$ \\

CFM-ECLD
& ECLD & 500
& $99.5 \pm 0.3$
& $99.5 \pm 0.2$
& N/A
& $0.305$
& $29.955 \pm 0.031$ & $33.84$ \\

\midrule
VAE Sampling & N/A & N/A & $93.1 \pm 0.7 $ & $98.45 \pm 0.2$ & N/A & $2.88$ & N/A& N/A \\
\midrule

Latent FM
& Heun & 2
& $96.7 \pm 0.6$
& $99.6 \pm 0.1$
& N/A
& $0.390$
& $1.840 \pm 0.005$ & $536.0$ \\

Latent FM
& Heun & 5
& $97.9 \pm 0.4$
& $\textbf{99.7} \pm 0.1$
& N/A
& $0.156$
& $1.836 \pm 0.007$ & $544.4$ \\

Latent FM
& Heun & 20
& $98.2 \pm 0.4$
& $\textbf{99.7} \pm 0.2$
& N/A
& $0.110$
& $1.870 \pm 0.014$ & $536.1$ \\

Latent FM
& Heun & 500
& $98.4 \pm 0.3$
& $\textbf{99.7} \pm 0.2$
& N/A
& $0.127$
& $3.095 \pm 0.019$ & $324.6$ \\
\midrule
Lat. FM + VG
& Midpoint & 20
& $99.5 \pm 0.2$
& $\textbf{99.7} \pm 0.1$
& N/A
& $0.110$
& $1.910 \pm 0.023$ & $531.8$ \\

VA-VFM
& Euler & 20
& $99.7 \pm 0.1$
& $99.3 \pm 0.2$
& N/A
& $0.760$
& $1.880 \pm 0.021$ & $539.3$ \\

\midrule
\textbf{PubChem16} & & & & & & & & $ $ \\
\midrule

DiGress
& Diffusion & 2
& $29.1 \pm 1.9$
& $99.4 \pm 0.5$
& $77.1 \pm 1.9$
& $10.44$
& $0.403 \pm 0.044$ & $566.7$ \\

DiGress
& Diffusion & 5
& $58.7 \pm 1.8$
& $99.9 \pm 0.1$
& $82.2 \pm 1.3$
& $6.060$
& $0.817 \pm 0.029$ & $604.2$ \\

DiGress
& Diffusion & 20
& $94.9 \pm 0.7$
& $100.0 \pm 0.0$
& $83.1 \pm 1.2$
& $0.945$
& $2.928 \pm 0.026$ & $275.8$ \\

DiGress
& Diffusion & 500
& $\textbf{99.6} \pm 0.2$
& $\textbf{100.0} \pm 0.0$
& $81.6 \pm 1.1$
& $0.356$
& $70.637 \pm 0.058$ & $11.8$ \\

\midrule

CFM-CSD
& CSD & 1
& $87.1 \pm 1.2$
& $99.6 \pm 0.3$
& $75.7 \pm 1.5$
& $6.978$
& $0.510 \pm 0.002$ & $1318.6$ \\

CFM-CSD
& CSD & 2
& $84.8 \pm 1.2$
& $\textbf{100.0} \pm 0.1$
& $87.8 \pm 1.1$
& $2.901$
& $0.670 \pm 0.002$ & $1137.9$ \\

CFM-CSD
& CSD & 5
& $92.7 \pm 1.0$
& $\textbf{100.0} \pm 0.0$
& $88.6 \pm 1.2$
& $1.768$
& $1.151 \pm 0.003$ & $730.7$ \\

CFM-CSD
& CSD & 20
& $97.0 \pm 0.7$
& $\textbf{100.0} \pm 0.0$
& $\textbf{89.2} \pm 1.0$
& $1.155$
& $3.562 \pm 0.004$ & $248.7$ \\

CFM-CSD
& CSD & 500
& $98.3 \pm 0.4$
& $99.9 \pm 0.1$
& $87.8 \pm 1.1$
& $0.70$
& $82.362 \pm 0.111$ & $10.71$ \\

\midrule

CFM-ECLD
& ECLD & 1
& $95.9 \pm 0.7$
& $98.7 \pm 0.4$
& $64.7 \pm 1.3$
& $7.034$
& $0.511 \pm 0.002$ & $1227.2$ \\

CFM-ECLD
& ECLD & 2
& $93.7 \pm 0.6$
& $\textbf{100.0} \pm 0.1$
& $86.0 \pm 1.0$
& $2.949$
& $0.670 \pm 0.002$ & $1231.6$ \\

CFM-ECLD
& ECLD & 5
& $96.1 \pm 0.5$
& $\textbf{100.0} \pm 0.0$
& $88.9 \pm 1.1$
& $2.339$
& $1.153 \pm 0.002$ & $758.8$ \\

CFM-ECLD
& ECLD & 20
& $97.4 \pm 0.5$
& $\textbf{100.0} \pm 0.0$
& $89.0 \pm 1.1$
& $2.083$
& $3.554 \pm 0.004$ & $249.8$ \\

CFM-ECLD
& ECLD & 500
& $97.8 \pm 0.4$
& $99.9 \pm 0.1$
& $88.5 \pm 1.1$
& $1.462$
& $82.361 \pm 0.105$ & $10.75$ \\

\midrule
VAE Sampling & N/A & N/A & $67.3 \pm 1.5 $ & $100.0 \pm 0.0$ & $91.7 \pm 1.0$ & $4.201$ & N/A& N/A \\
\midrule

Latent FM
& Heun & 2
& $94.7 \pm 0.6$
& $99.8\pm 0.1$
& $52.8 \pm 1.6$
& $2.33$
& $ 3.498\pm 0.030$ & $146.1$ \\

Latent FM
& Heun & 5
& $95.6 \pm 0.5$
& $100.0\pm 0.0$
& $79.0 \pm 1.4$
& $0.409$
& $3.757 \pm 0.027$ & $205.8$ \\

Latent FM
& Heun & 20
& $97.3 \pm 0.5$
& $\textbf{100.0} \pm 0.0$
& $81.2 \pm 1.3$
& $0.190$
& $4.260 \pm 0.039$ & $189.9$ \\

Latent FM
& Heun & 500
& $97.7 \pm 0.5$
& $\textbf{100.0} \pm 0.0$
& $80.4 \pm 1.5$
& $\textbf{0.171}$
& $10.274 \pm 0.056$ & $78.3$ \\
\midrule
Lat. FM + VG
& Midpoint & 20
& $97.8 \pm 0.4$
& $\textbf{100.0} \pm 0.0$
& $81.0 \pm 1.1$
& $0.210$
& $4.430 \pm 0.040$ & $183.1$ \\

VA-VFM
& Euler & 20
& $98.6 \pm 0.3$
& $99.9 \pm 0.1$
& $82.1 \pm 1.2$
& $1.370$
& $4.420 \pm 0.029$ & $187.4$ \\

\midrule
\textbf{PubChem32}  & & & & & & & & $ $ \\
\midrule

DiGress & Diffusion & 20  & $89.2\pm0.7$ & $\textbf{100.0}\pm0.0$ & $93.0\pm0.8$ & $3.24$ & $7.922 \pm 0.133$ & $107.23$ \\
DiGress & Diffusion & 500 & $99.0\pm0.3$ & $\textbf{100.0}\pm0.0$ & $95.1\pm0.5$ & $1.17$ & $193.77 \pm 0.054$ & $3.406$ \\

\midrule
VAE Sampling & N/A & N/A & $11.1 \pm 0.1$ & $100.0 \pm 0.0$ & $92.61 \pm 0.02 $ & $11.33$ & N/A & N/A \\
\midrule
Latent FM            & Heun     & 2   & $78.6 \pm 0.8$ & $93.7 \pm 0.9 $ & $36.7 \pm 1.4$ & $13.52$ & $4.764 \pm 0.031$ & $58.1$ \\
Latent FM            & Heun     & 20  & $74.5 \pm 0.6$ & $\textbf{100.0}\pm 0.0$ & $91.1\pm 1.1$ & $0.868$ & $9.425 \pm 0.123$ & $73.8$ \\
Latent FM            & Heun     & 500 & $74.4\pm1.9$ & $\textbf{100.0}\pm 0.0$ & $91.2\pm1.2$ & $\textbf{0.856}$ & $17.927 \pm 0.126$ & $38.8$ \\

\bottomrule
\end{tabular}}
\end{table}

We first evaluate whether graph generation can be performed effectively by modeling the distribution of compact graph-level continuous representations, rather than explicitly generating discrete node and edge variables. This comparison is designed to evaluate not only generation quality, but also the computational trade-offs associated with the different generative representations. 

\FloatBarrier

DiGress operates explicitly in discrete graph space through a long denoising trajectory, while CFM is specifically designed for efficient few-step categorical generation. We report the standard metrics of validity, uniqueness, novelty, Fréchet Chemnet Distance~\citep{preuer2018fcd} to assess distributional fidelity, and end-to-end sampling time. We additionally report computational efficiency to compare latent and explicit graph generation beyond sample quality. Finally, we compute a metric, which we call \textbf{VUN/T}, which is computed as the quadruple product of validity, uniqueness, novelty and batch size, divided by the generation time; it corresponds to the number of valid, unique and novel molecules obtained per second, and serves as a summary of the generation efficiency. All results are reported in Table \ref{tab:uncond}.

Explicit methods dominate on QM9, composed of very small molecules, although all methods achieve high performance on this dataset. Already at this scale, a key advantage of latent flow matching emerges: total generation time grows affinely with the number of integration steps for both approaches, but with a much shallower slope for latent FM, and the dominant cost is the single, fixed final decoding step. Explicit approaches, by contrast, pay a substantially steeper per-step cost, as every integration step operates on the full graph representation.

On PubChem16, our model shows a clear advantage in distributional fidelity, achieving substantially better FCD than both CFM variants, even at only 20 NFE, and outperforming DiGress by a significant margin. Notably, increasing the number of integration steps has only a marginal effect on our model's fidelity, in contrast to explicit methods. DiGress attains the highest validity rate, but only in its 500-step regime, which requires 70 seconds per batch, compared to 4 seconds for our model at 20 NFE (or 10 seconds at 500 NFE), while our model still achieves significantly lower (better) FCD in both regimes, with a \textbf{VUN/T} up to $17$ times higher.

On Pubchem32, we did not train CFM due to its very large training cost (73 hours on PubChem16, which is 7 times smaller). We therefore compare to DiGress only. First of all we notice that, like for the other datasets, our model has significantly improved the generation performance with regards to the base VAE, improving validity 7-fold. Nonetheless, it does not manage to fully compensate for this, stagnating around $74.5\%$ validity while Digress reaches $99\%$ with full $500$ steps.

Validity guidance further improves the latent FM by increasing chemical validity while preserving its distributional fidelity. The validity-aware VFM increases validity on QM9 and PubChem16, but at the cost of significant FCD; it should therefore be viewed as a specialized alternative for applications in which validity is the primary objective. On PubChem32, validity guidance fails at improving the validity on that dataset, which we attribute to the complexity of the latent space for that dataset relative to the small size of the validity predictor's training set (800,000 samples).

\paragraph{Training efficiency.}

\begin{table}[t]
\centering
\caption{
Single-GPU training cost on QM9 and PubChem16. Times for the proposed methods
assume a fixed pretrained graph representation and therefore exclude the
one-time encoder-decoder pretraining cost.
}
\label{tab:training_times}
\begin{tabular}{lcc}
\toprule
\textbf{Method / component} &
\textbf{QM9} &
\textbf{PubChem16} \\
\midrule

DiGress
& 7 h 34 min
& 16 h 55 min \\

CFM-CSD
& 63 h 49 min
& 73 h 21 min \\

CFM-ECLD
& 70 h 53 min
& 71 h 40 min \\

\midrule

Latent FM
& 27 min
& 3 h 50 min \\

Validity dataset construction
& 2 h 06 min
& 8 h 05 min \\

Validity predictor
& 1 min 28 s
& 1 min 12 s \\

Validity-aware VFM training
& 15 min 24 s
& 12 h 08 min \\

\midrule

Latent FM + validity guidance
& 2 h 35 min
& 11 h 57 min \\

Complete latent pipeline
& 2 h 50 min
& 24 h 05 min \\

\bottomrule
\end{tabular}
\end{table}

The computational profiles also differ substantially at training time, as reported in Tables~\ref{tab:training_times}. CFM provides the fastest inference but requires the longest training among the evaluated approaches. DiGress has a more moderate training cost, while its strongest generation results rely on long sampling trajectories. In contrast, latent flow matching is particularly inexpensive to train, as it exploits a pre-trained VAE, with a training cost on QM9 that is roughly two orders of magnitude below that of CFM.

Even when validity dataset construction, classifier training, and the validity-aware endpoint model are included, the complete latent pipeline remains substantially cheaper to train than CFM on QM9. A similar trend is observed on PubChem16, where the latent FM alone remains considerably less expensive to train than the explicit categorical flow models (Table~\ref{tab:training_times}). These comparisons exclude the one-time cost of learning the graph encoder-decoder used by the latent models. However, this representation is shared across downstream tasks rather than being trained specifically for each generative objective. Once learned, the same continuous graph space is reused for unconditional generation, validity guidance, validity-aware generation, and property control. Even if trained specifically for latent flow matching purposes, the cost of representation learning can therefore be amortized across multiple tasks, which is a central advantage of decoupling graph representation learning from generative modeling.

\subsection{Property-Guided Generation}
\label{sec:property_results}

We study independent control of two complementary molecular properties, molecular weight (MW) and logP, prescribing one property at a time. MW mainly reflects molecular size and atomic composition, whereas logP depends more strongly on molecular structure and functional-group arrangement. This allows us to assess whether a single continuous latent representation can support distinct forms of molecular control rather than a single property class.

\paragraph{Independent property control.}

For each property, we condition the latent Flow Matching model on the target and optionally refine sampling with gradient-based guidance from a property predictor. Targets cover both high-density and more extreme regions of the training distribution, allowing us to assess controllability beyond frequent property values.

\begin{figure}[t]
\centering

\begin{subfigure}{0.48\linewidth}
    \centering
    \includegraphics[width=\linewidth]
    {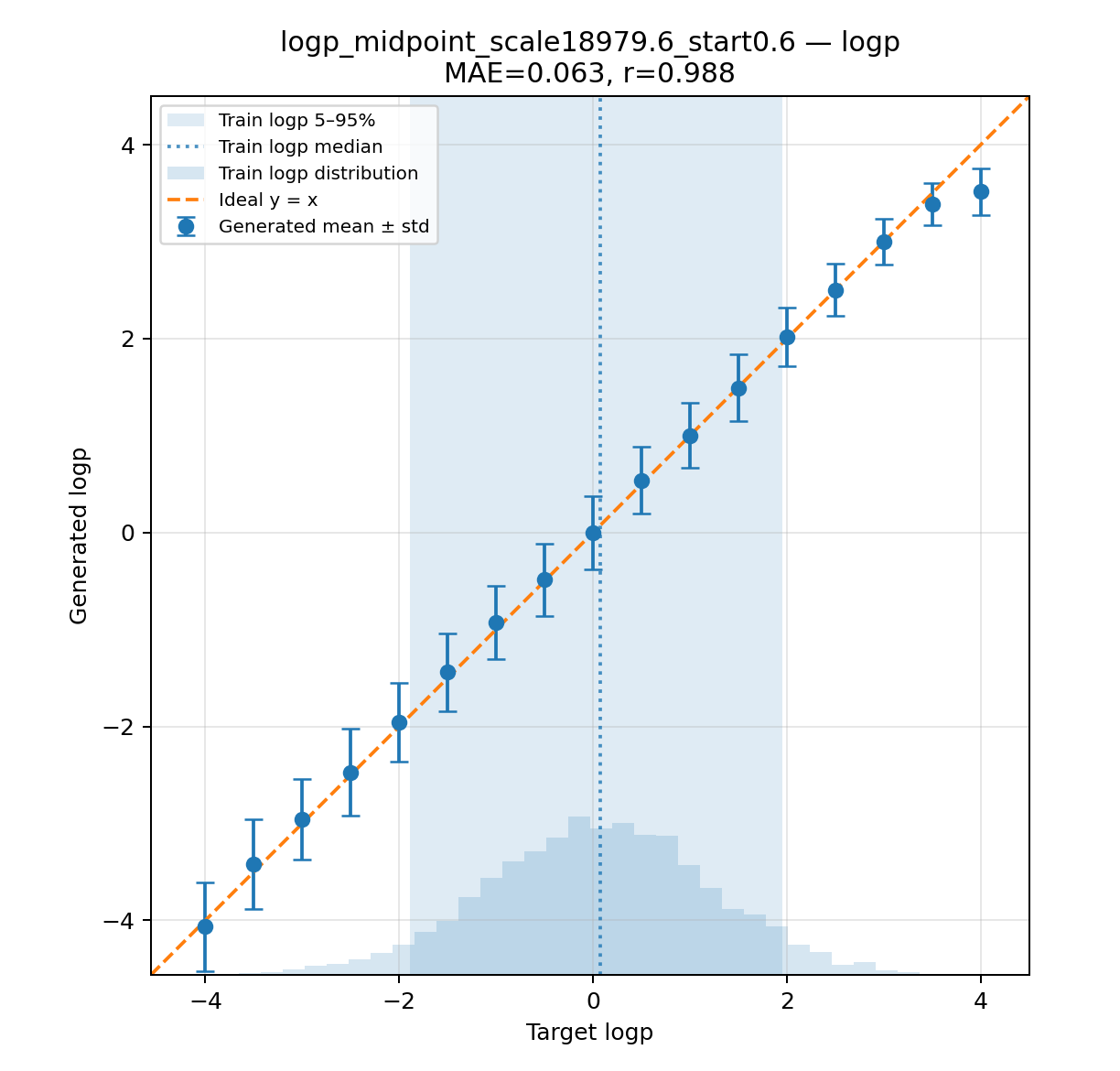}
    \caption{QM9 - logP.}
\end{subfigure}
\hfill
\begin{subfigure}{0.48\linewidth}
    \centering
    \includegraphics[width=\linewidth]
    {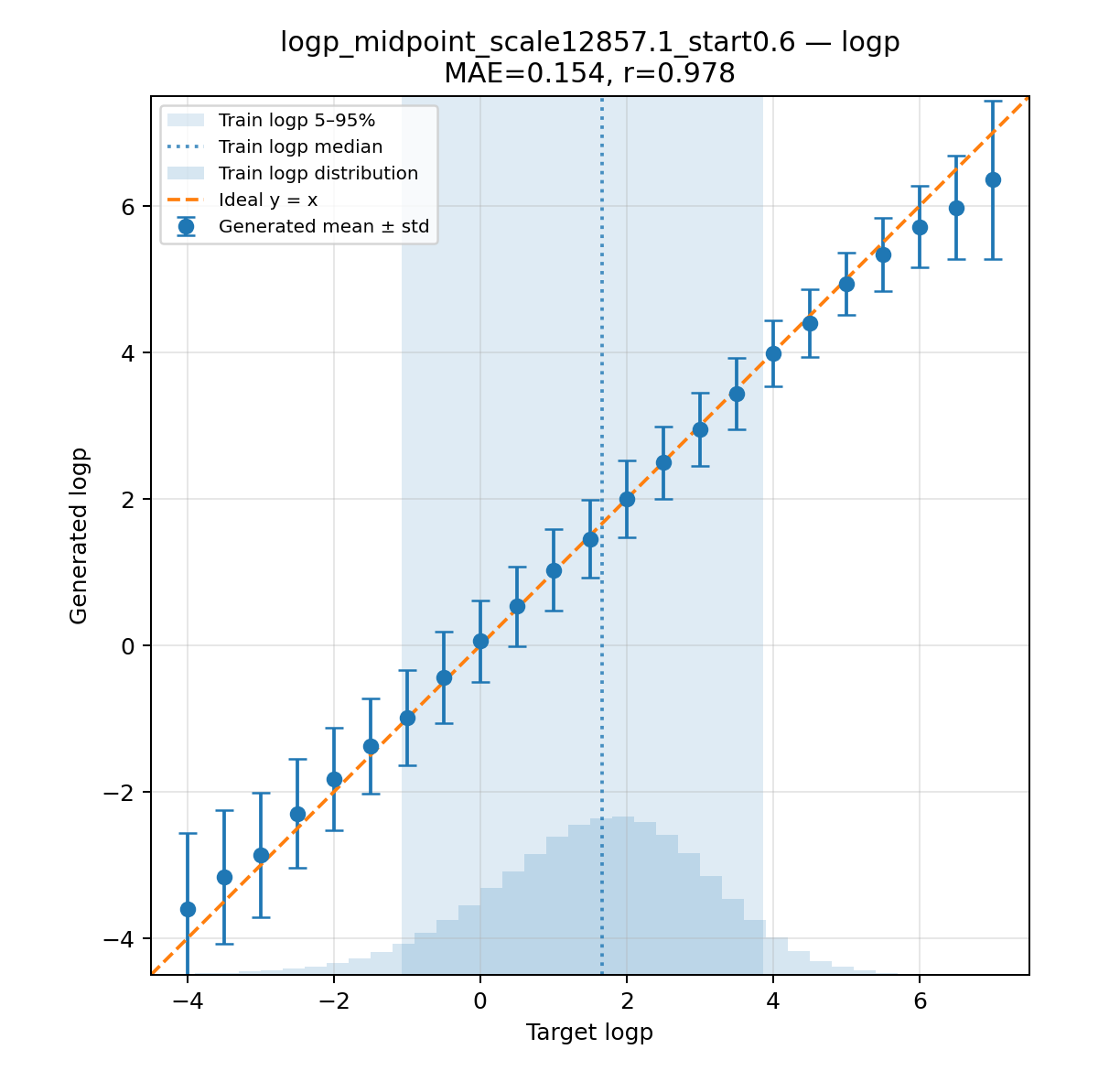}
    \caption{PubChem16 - logP.}
\end{subfigure}

\vspace{0.3cm}

\begin{subfigure}{0.48\linewidth}
    \centering
    \includegraphics[width=\linewidth]
    {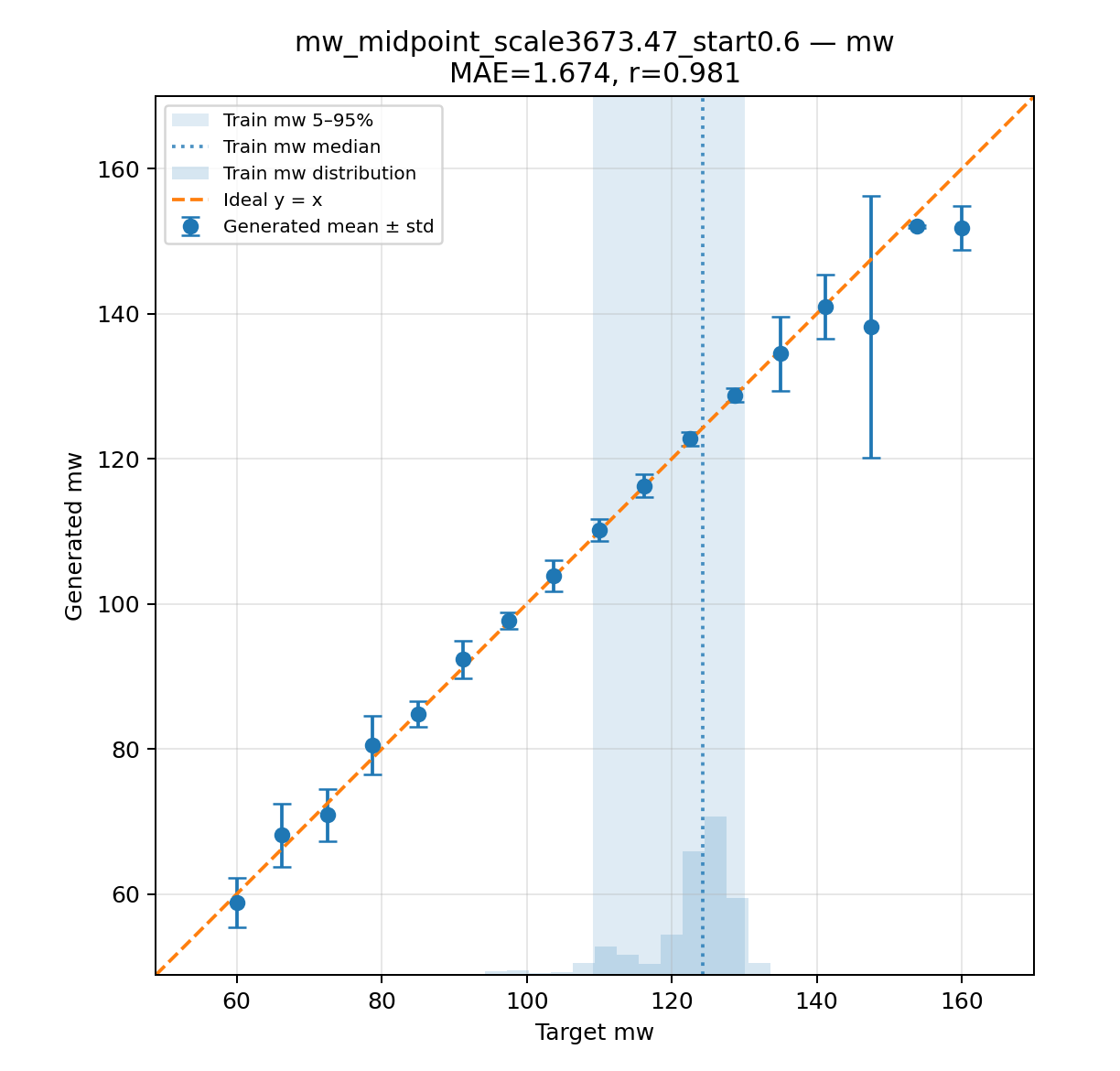}
    \caption{QM9 - molecular weight.}
\end{subfigure}
\hfill
\begin{subfigure}{0.48\linewidth}
    \centering
    \includegraphics[width=\linewidth]
    {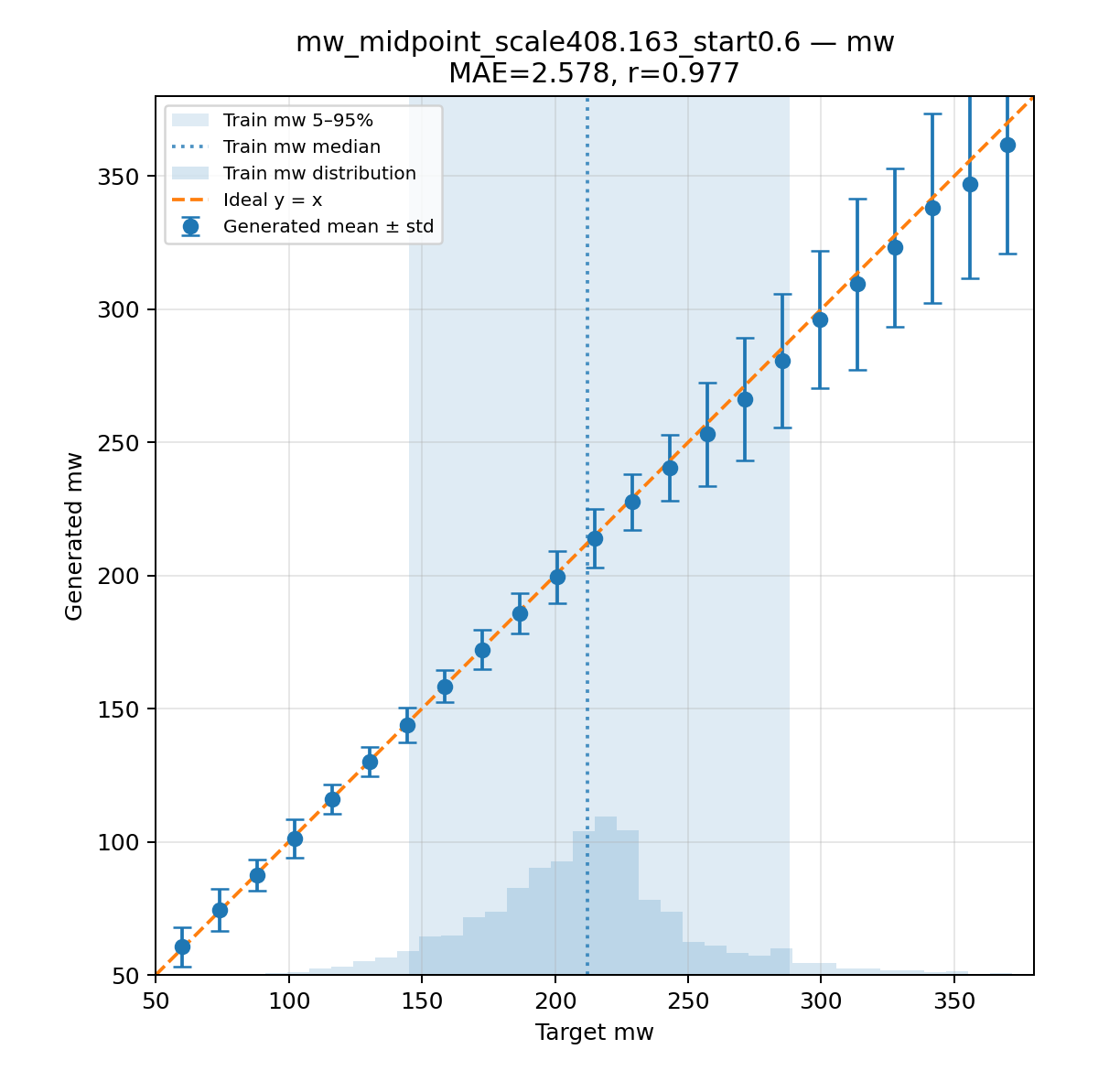}
    \caption{PubChem16 - molecular weight.}
\end{subfigure}

\caption{
Independent property-controlled generation on QM9 and PubChem16.
Points show the mean property of generated molecules for each requested
target, with error bars indicating one standard deviation. The dashed line
corresponds to ideal control, while the shaded region indicates the
5-95\% range of the training distribution.
}
\label{fig:property_guidance}
\end{figure}

Figure~\ref{fig:property_guidance} compares requested and generated properties for the best guided configurations. Across QM9 and PubChem16, generated means closely track the targets over a broad range, showing that the same latent representation supports both unconditional generation and property control. Qualitative examples are provided in Appendix~\ref{app:property_guided_examples}.

Although conditioning alone already provides effective property control, gradient-based guidance proves substantially more effective at improving target fidelity, with the largest gains observed for logP and consistent improvements for MW (Table~\ref{tab:property_guidance_ablation}).

\begin{table}[t]
\centering
\caption{
Effect of gradient-based property guidance on target fidelity.
We compare conditional flow matching (CFM) with CFM
augmented with property-predictor guidance.
Lower MAE and higher Pearson correlation $r$ indicate better property control.
}
\label{tab:property_guidance_ablation}
\begin{tabular}{llcccc}
\toprule
\textbf{Dataset} &
\textbf{Property} &
\multicolumn{2}{c}{\textbf{Conditional FM}} &
\multicolumn{2}{c}{\textbf{+ Guidance}} \\
\cmidrule(lr){3-4}
\cmidrule(lr){5-6}
&
&
\textbf{MAE} $\downarrow$ &
$\mathbf{r}$ $\uparrow$ &
\textbf{MAE} $\downarrow$ &
$\mathbf{r}$ $\uparrow$ \\
\midrule

QM9
& logP
& 0.100
& 0.978
& \textbf{0.063}
& \textbf{0.988} \\

QM9
& MW
& 1.705
& 0.979
& \textbf{1.674}
& \textbf{0.981} \\

\midrule

PubChem16
& logP
& 0.282
& 0.966
& \textbf{0.154}
& \textbf{0.978} \\

PubChem16
& MW
& 3.040
& 0.976
& \textbf{2.578}
& \textbf{0.977} \\

\bottomrule
\end{tabular}
\end{table}

\paragraph{Influence of the latent representation.}
Both deterministic and variational GraViti representations support effective
property control, although the variational latent space consistently yields
more accurate target matching (Appendix~\ref{app:latent_representation_guidance}).

\paragraph{Validity under property control.}

Target fidelity alone is insufficient to characterize controlled generation:
a model may reach increasingly extreme property values by moving toward latent
regions that decode poorly. We therefore distinguish between {global}
and {local} validity. Global validity evaluates the conditional generation regime in which target
properties are sampled according to their empirical training distribution. It
therefore gives greater weight to targets lying in high-density regions of the
property space. Local validity instead evaluates validity separately at each
prescribed target and reveals how generation quality evolves when moving toward
rare or out-of-distribution property values.

Under global conditioning, property guidance preserves high validity and uniqueness on both datasets, remaining close to the unconditional regime on QM9 and above $0.92$ validity with maximal uniqueness on PubChem16. Detailed results are reported in Appendix~\ref{app:global_val}.

Global validity, however, can conceal substantial variations across the property range. Since targets are sampled according to the empirical training distribution, the resulting average is dominated by high-density regions and provides limited information about the behavior of the model near the boundaries of, or beyond, the training distribution. We therefore additionally report \emph{local validity}, computed independently for each prescribed target value. Local validity and uniqueness remain high within the training distribution, while target-wise results in Appendix~\ref{app:local_validity} show that validity degrades primarily for increasingly rare or out-of-distribution targets. These results also qualitatively indicate that our models remain reasonably robust beyond the training range.

\section{Conclusion}

In this paper, we introduced a method for molecule generation using a flow matching algorithm in the latent space of a high-fidelity graph-level autoencoder, leveraging on the recent progress on whole-molecule representations. We showed that our approach reaches state-of-the-art distributional fidelity (measured as the FCD), even when set up against more complex  explicit graph generation baselines. While a decrease in validity is observed compared to DiGress (up to 25 points on PubChem32), the efficiency makes up for that by allowing to generate up to 10 times faster while retaining the larger distributional fidelity. We also demonstrate that our approach integrates guided generation seamlessly, leveraging the same latent representation and providing highly fidel results even outside of the dataset's distribution ranges.

Overall, this work offers a promising approach to graph and molecule generation processes that scale better, avoiding the high cost of in-graph-space explicit propagation. Future work could include structuring the latent space with explicit chemical properties in mind to further push the performance and application-suitability of the generated molecules.
\clearpage 

\bibliographystyle{unsrt}
\bibliography{bib}

@inproceedings{lipman2022flow,
  title     = {Flow Matching for Generative Modeling},
  author    = {Lipman, Yaron and Chen, Ricky T. Q. and Ben-Hamu, Heli
               and Nickel, Maximilian and Le, Matt},
  booktitle = {International Conference on Learning Representations},
  year      = {2022},
  url       = {https://arxiv.org/abs/2210.02747}
}

@inproceedings{vignac2023digress,
  title     = {{DiGress}: Discrete Denoising Diffusion for Graph Generation},
  author    = {Vignac, Clement and Krawczuk, Igor and Siraudin, Antoine
               and Wang, Bohan and Cevher, Volkan and Frossard, Pascal},
  booktitle = {International Conference on Learning Representations},
  year      = {2023},
  url       = {https://arxiv.org/abs/2209.14734}
}

@inproceedings{eijkelboom2024vfm,
  title     = {Variational Flow Matching for Graph Generation},
  author    = {Eijkelboom, Floor and Bartosh, Grigory
               and Naesseth, Christian Andersson and Welling, Max
               and van de Meent, Jan-Willem},
  booktitle = {Advances in Neural Information Processing Systems},
  year      = {2024},
  url       = {https://arxiv.org/abs/2406.04843}
}

@inproceedings{
li2025towards,
title={Towards Understanding the Mechanisms of Classifier-Free Guidance},
author={Xiang Li and Rongrong Wang and Qing Qu},
booktitle={The Thirty-ninth Annual Conference on Neural Information Processing Systems},
year={2025},
url={https://openreview.net/forum?id=bRAm7A02Qm}
}

@InProceedings{perez2018film,
  title={FiLM: Visual Reasoning with a General Conditioning Layer},
  author={Ethan Perez and Florian Strub and Harm de Vries and Vincent Dumoulin and Aaron C. Courville},
  booktitle={AAAI},
  year={2018}
}

@inproceedings{pigvae,
 author = {Winter, Robin and Noe, Frank and Clevert, Djork-Arn\'{e}},
 booktitle = {Advances in Neural Information Processing Systems},
 editor = {M. Ranzato and A. Beygelzimer and Y. Dauphin and P.S. Liang and J. Wortman Vaughan},
 pages = {9559--9573},
 publisher = {Curran Associates, Inc.},
 title = {Permutation-Invariant Variational Autoencoder for Graph-Level Representation Learning},
 url = {https://proceedings.neurips.cc/paper_files/paper/2021/file/4f3d7d38d24b740c95da2b03dc3a2333-Paper.pdf},
 volume = {34},
 year = {2021}
}

@article{roos2026categorical,
  title         = {Categorical Flow Maps},
  author        = {Roos, Daan and Davis, Oscar and Eijkelboom, Floor
                   and Bronstein, Michael and Welling, Max
                   and Ceylan, Ismail Ilkan and Ambrogioni, Luca
                   and van de Meent, Jan-Willem},
  journal       = {arXiv preprint arXiv:2602.12233},
  year          = {2026},
  eprint        = {2602.12233},
  archivePrefix = {arXiv},
  primaryClass  = {cs.LG},
  url           = {https://arxiv.org/abs/2602.12233}
}

@article{bresson2026graviti,
  title         = {{GraViti}: Graph-Level Variational Autoencoders with Relaxed
                   Permutation Invariance},
  author        = {Bresson, Roman and Divriotis, Konstantinos
                   and Lutzeyer, Johannes F. and Evdaimon, Iakovos
                   and Vazirgiannis, Michalis},
  journal       = {arXiv preprint arXiv:2605.16668},
  year          = {2026},
  eprint        = {2605.16668},
  archivePrefix = {arXiv},
  primaryClass  = {cs.LG},
  url           = {https://arxiv.org/abs/2605.16668}
}

@article{gomez2018automatic,
  title   = {Automatic Chemical Design Using a Data-Driven Continuous
             Representation of Molecules},
  author  = {G{\'o}mez-Bombarelli, Rafael and Wei, Jennifer N.
             and Duvenaud, David and Hern{\'a}ndez-Lobato, Jos{\'e} Miguel
             and S{\'a}nchez-Lengeling, Benjam{\'i}n and Sheberla, Dennis
             and Aguilera-Iparraguirre, Jorge and Hirzel, Timothy D.
             and Adams, Ryan P. and Aspuru-Guzik, Al{\'a}n},
  journal = {ACS Central Science},
  volume  = {4},
  number  = {2},
  pages   = {268--276},
  year    = {2018},
  doi     = {10.1021/acscentsci.7b00572}
}

@inproceedings{wei2024chemflow,
  title     = {Navigating Chemical Space with Latent Flows},
  author    = {Wei, Guanghao and Huang, Yining and Duan, Chenru
               and Song, Yue and Du, Yuanqi},
  booktitle = {Advances in Neural Information Processing Systems},
  year      = {2024},
  url       = {https://arxiv.org/abs/2405.03987}
}

@inproceedings{dhariwal2021diffusion,
  title     = {Diffusion Models Beat {GANs} on Image Synthesis},
  author    = {Dhariwal, Prafulla and Nichol, Alexander},
  booktitle = {Advances in Neural Information Processing Systems},
  volume    = {34},
  year      = {2021}
}

@article{preuer2018fcd,
  title   = {Fr{\'e}chet ChemNet Distance: A Metric for Generative Models
             for Molecules in Drug Discovery},
  author  = {Preuer, Kristina and Renz, Philipp and Unterthiner, Thomas
             and Hochreiter, Sepp and Klambauer, G{\"u}nter},
  journal = {Journal of Chemical Information and Modeling},
  volume  = {58},
  number  = {9},
  pages   = {1736--1741},
  year    = {2018},
  doi     = {10.1021/acs.jcim.8b00234}
}

@article{ramakrishnan2014qm9,
  title   = {Quantum Chemistry Structures and Properties of 134 Kilo Molecules},
  author  = {Ramakrishnan, Raghunathan and Dral, Pavlo O.
             and Rupp, Matthias and von Lilienfeld, O. Anatole},
  journal = {Scientific Data},
  volume  = {1},
  pages   = {140022},
  year    = {2014},
  doi     = {10.1038/sdata.2014.22}
}

@article{tong2024improving,
  title={Improving and Generalizing Flow-Based Generative Models with Minibatch Optimal Transport},
  author={Tong, Alexander and Fatras, Kilian and Malkin, Nikolay and Huguet, Guillaume and Zhang, Yanlei and Rector-Brooks, Jarrid and Wolf, Guy and Bengio, Yoshua},
  journal={Transactions on Machine Learning Research},
  year={2024}
}

@inproceedings{
zhou2024unifying,
title={Unifying Generation and Prediction on Graphs with Latent Graph Diffusion},
author={Cai Zhou and Xiyuan Wang and Muhan Zhang},
booktitle={The Thirty-eighth Annual Conference on Neural Information Processing Systems},
year={2024},
url={https://openreview.net/forum?id=lvibangnAs}
}

@ARTICLE{10508504,
  author={Yang, Ling and Huang, Zhilin and Zhang, Zhilong and Liu, Zhongyi and Hong, Shenda and Zhang, Wentao and Yang, Wenming and Cui, Bin and Zhang, Luxia},
  journal={IEEE Transactions on Knowledge and Data Engineering}, 
  title={Graphusion: Latent Diffusion for Graph Generation}, 
  year={2024},
  volume={36},
  number={11},
  pages={6358-6369},
  doi={10.1109/TKDE.2024.3389783}}

@misc{lipman2024flowmatchingguidecode,
      title={Flow Matching Guide and Code}, 
      author={Yaron Lipman and Marton Havasi and Peter Holderrieth and Neta Shaul and Matt Le and Brian Karrer and Ricky T. Q. Chen and David Lopez-Paz and Heli Ben-Hamu and Itai Gat},
      year={2024},
      eprint={2412.06264},
      archivePrefix={arXiv},
      primaryClass={cs.LG},
      url={https://arxiv.org/abs/2412.06264}, 
}

@inproceedings{gen3,
 author = {Zahirnia, Kiarash and Schulte, Oliver and Naddaf, Parmis and Li, Ke},
 booktitle = {Advances in Neural Information Processing Systems},
 doi = {10.52202/068431-2200},
 editor = {S. Koyejo and S. Mohamed and A. Agarwal and D. Belgrave and K. Cho and A. Oh},
 pages = {30347--30361},
 publisher = {Curran Associates, Inc.},
 title = {Micro and Macro Level Graph Modeling for Graph Variational Auto-Encoders},
 url = {https://proceedings.neurips.cc/paper_files/paper/2022/file/c400474e8a36d0812fdee52739288b12-Paper-Conference.pdf},
 volume = {35},
 year = {2022}
}

@inproceedings{qinmadeira2024defog,
  title     = {DeFoG: Discrete Flow Matching for Graph Generation},
  author    = {Qin, Yiming and Madeira, Manuel and Thanou, Dorina and Frossard, Pascal},
  booktitle = {International Conference on Machine Learning (ICML)},
  year      = {2025},
}

@article{kipf_variational_2016,
  author = {Kipf, Thomas N. and Welling, Max},
  journal = {arXiv:1611.07308 [cs, stat]},
  language = {en},
  month = nov,
  note = {arXiv: 1611.07308},
  title = {Variational {Graph} {Auto}-{Encoders}},
  url = {http://arxiv.org/abs/1611.07308},
  urldate = {2020-02-02},
  year = 2016
}

@inproceedings{
grale,
title={The quest for the {GRA}ph Level autoEncoder ({GRALE})},
author={Paul Krzakala and Gabriel Melo and Charlotte Laclau and Florence d'Alch{\'e}-Buc and R{\'e}mi Flamary},
booktitle={The Thirty-ninth Annual Conference on Neural Information Processing Systems},
year={2025},
}

@ARTICLE{autoencoders_basic,
  author={Bengio, Yoshua and Courville, Aaron and Vincent, Pascal},
  journal={IEEE Transactions on Pattern Analysis and Machine Intelligence}, 
  title={Representation Learning: A Review and New Perspectives}, 
  year={2013},
  volume={35},
  number={8},
  pages={1798-1828},
  doi={10.1109/TPAMI.2013.50}}

@inproceedings{autoencodingvariational,
  author = {Kingma, Diederik P. and Welling, Max},
  booktitle = {2nd International Conference on Learning Representations, {ICLR} 2014, Banff, AB, Canada, April 14-16, 2014, Conference Track Proceedings},
  eprint = {http://arxiv.org/abs/1312.6114v10},
  eprintclass = {stat.ML},
  eprinttype = {arXiv},
  title = {{Auto-Encoding Variational Bayes}},
  year = 2014
}

@article{siraudin2026principled,
  title   = {Principled Latent Diffusion for Graphs via Laplacian Autoencoders},
  author  = {Siraudin, Antoine and Morris, Christopher},
  journal = {arXiv preprint arXiv:2601.13780},
  year    = {2026},
  note    = {Preprint},
  url     = {https://arxiv.org/abs/2601.13780}
}

@inproceedings{xu2023geometric,
  title={Geometric Latent Diffusion Models for 3D Molecule Generation},
  author={Minkai Xu and Alexander Powers and Ron Dror and Stefano Ermon and Jure Leskovec},
  booktitle={International Conference on Machine Learning},
  year={2023},
  organization={PMLR}
}

@InProceedings{juncvae,
  title = 	 {Junction Tree Variational Autoencoder for Molecular Graph Generation},
  author =       {Jin, Wengong and Barzilay, Regina and Jaakkola, Tommi},
  booktitle = 	 {Proceedings of the 35th International Conference on Machine Learning},
  pages = 	 {2323--2332},
  year = 	 {2018},
  editor = 	 {Dy, Jennifer and Krause, Andreas},
  volume = 	 {80},
  series = 	 {Proceedings of Machine Learning Research},
  month = 	 {10--15 Jul},
  publisher =    {PMLR},
  url = 	 {https://proceedings.mlr.press/v80/jin18a.html}
}

@inproceedings{Grammar,
author = {Kusner, Matt J. and Paige, Brooks and Hern\'{a}ndez-Lobato, Jos\'{e} Miguel},
title = {Grammar variational autoencoder},
year = {2017},
publisher = {JMLR.org},
booktitle = {Proceedings of the 34th International Conference on Machine Learning - Volume 70},
pages = {1945–1954},
numpages = {10},
location = {Sydney, NSW, Australia},
series = {ICML'17}
}

@article{qm9,
  title = {Quantum chemistry structures and properties of 134 kilo molecules},
  volume = {1},
  ISSN = {2052-4463},
  url = {http://dx.doi.org/10.1038/sdata.2014.22},
  DOI = {10.1038/sdata.2014.22},
  number = {1},
  journal = {Scientific Data},
  publisher = {Springer Science and Business Media LLC},
  author = {Ramakrishnan,  Raghunathan and Dral,  Pavlo O. and Rupp,  Matthias and von Lilienfeld,  O. Anatole},
  year = {2014},
  month = Aug 
}

@article{PubChem,
    author = {Kim, Sunghwan and Thiessen, Paul A. and Bolton, Evan E. and Chen, Jie and Fu, Gang and Gindulyte, Asta and Han, Lianyi and He, Jane and He, Siqian and Shoemaker, Benjamin A. and Wang, Jiyao and Yu, Bo and Zhang, Jian and Bryant, Stephen H.},
    title = {PubChem Substance and Compound databases},
    journal = {Nucleic Acids Research},
    volume = {44},
    number = {D1},
    pages = {D1202-D1213},
    year = {2016},
    month = {01},
    issn = {0305-1048},
    doi = {10.1093/nar/gkv951},
    url = {https://doi.org/10.1093/nar/gkv951},
    eprint = {https://academic.oup.com/nar/article-pdf/44/D1/D1202/9484096/gkv951.pdf},
}

@article{NGUYEN20253867,
title = {Transformer graph variational autoencoder for generative molecular design},
journal = {Biophysical Journal},
volume = {124},
number = {22},
pages = {3867-3875},
year = {2025},
issn = {0006-3495},
doi = {https://doi.org/10.1016/j.bpj.2025.01.022},
url = {https://www.sciencedirect.com/science/article/pii/S0006349525000359},
author = {Trieu Nguyen and Aleksandra Karolak}
}

@ARTICLE{Hu2023-xz,
  title     = "{ScaffoldGVAE}: scaffold generation and hopping of drug
               molecules via a variational autoencoder based on multi-view
               graph neural networks",
  author    = "Hu, Chao and Li, Song and Yang, Chenxing and Chen, Jun and
               Xiong, Yi and Fan, Guisheng and Liu, Hao and Hong, Liang",
  journal   = "J. Cheminform.",
  publisher = "Springer Science and Business Media LLC",
  volume    =  15,
  number    =  1,
  pages     = "91",
  month     =  oct,
  year      =  2023,
  copyright = "https://creativecommons.org/licenses/by/4.0",
  language  = "en"
}

@article{evdaimon2024neural,
  title={Neural Graph Generator: Feature-Conditioned Graph Generation using Latent Diffusion Models},
  author={Evdaimon, Iakovos and Nikolentzos, Giannis and Xypolopoulos, Christos and Kammoun, Ahmed and Chatzianastasis, Michail and Abdine, Hadi and Vazirgiannis, Michalis},
  journal={arXiv preprint arXiv:2403.01535},
  year={2024}
}

\clearpage
\appendix
\section{Influence of the Latent Representation on Property Control}
\label{app:latent_representation_guidance}

To further assess the influence of the graph-level representation on property-controlled generation, we compare the deterministic and variational GraViti latent spaces on PubChem16. In both cases, we use the same dataset, controlled properties, and property-guidance protocol, changing only the latent representation in which the generative dynamics are learned.

\begin{figure}[!h]
    \centering

    \begin{minipage}[t]{0.48\linewidth}
        \centering
        \includegraphics[width=\linewidth]
        {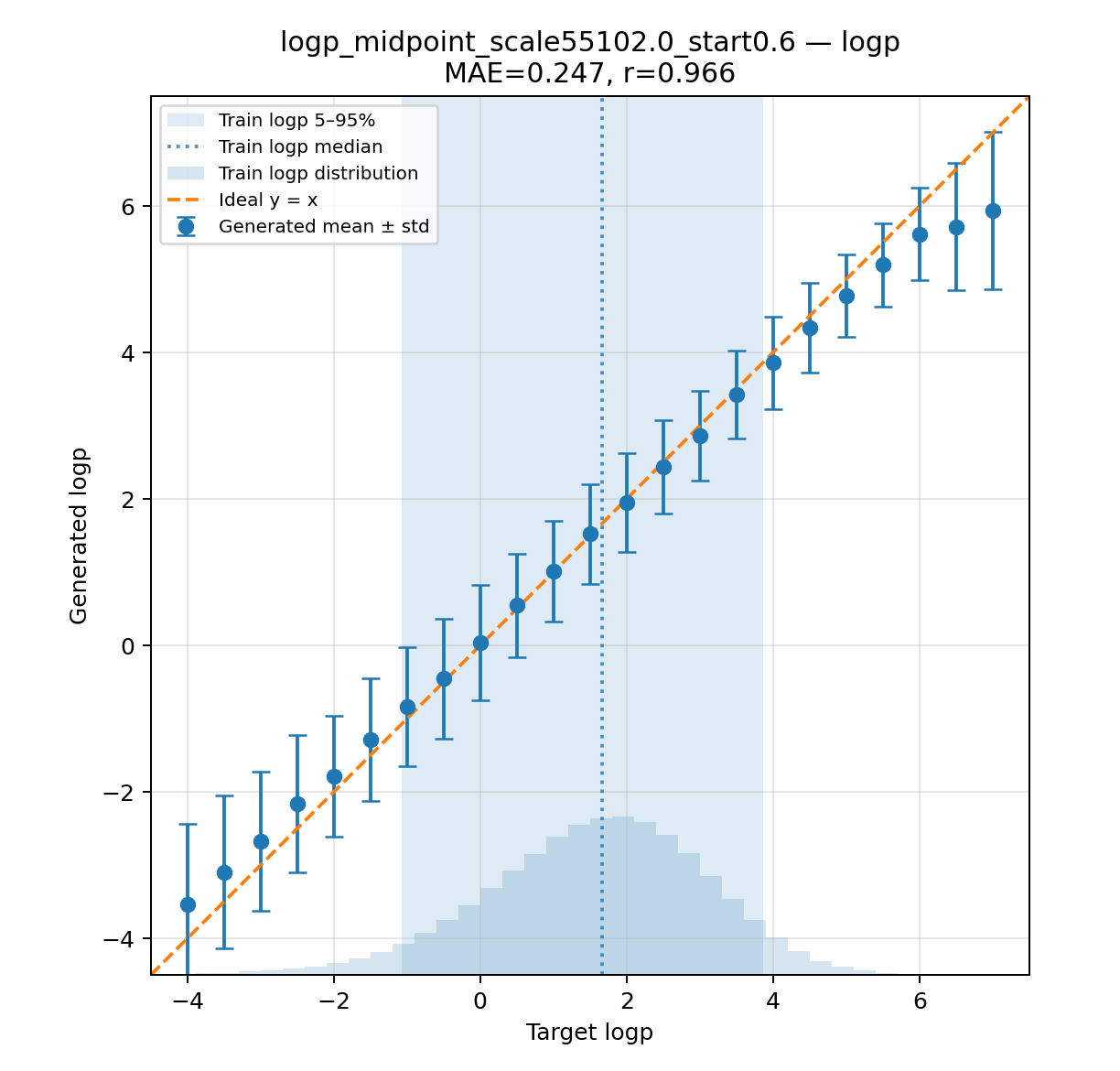}
        \caption*{(a) logP}
    \end{minipage}
    \hfill
    \begin{minipage}[t]{0.48\linewidth}
        \centering
        \includegraphics[width=\linewidth]
        {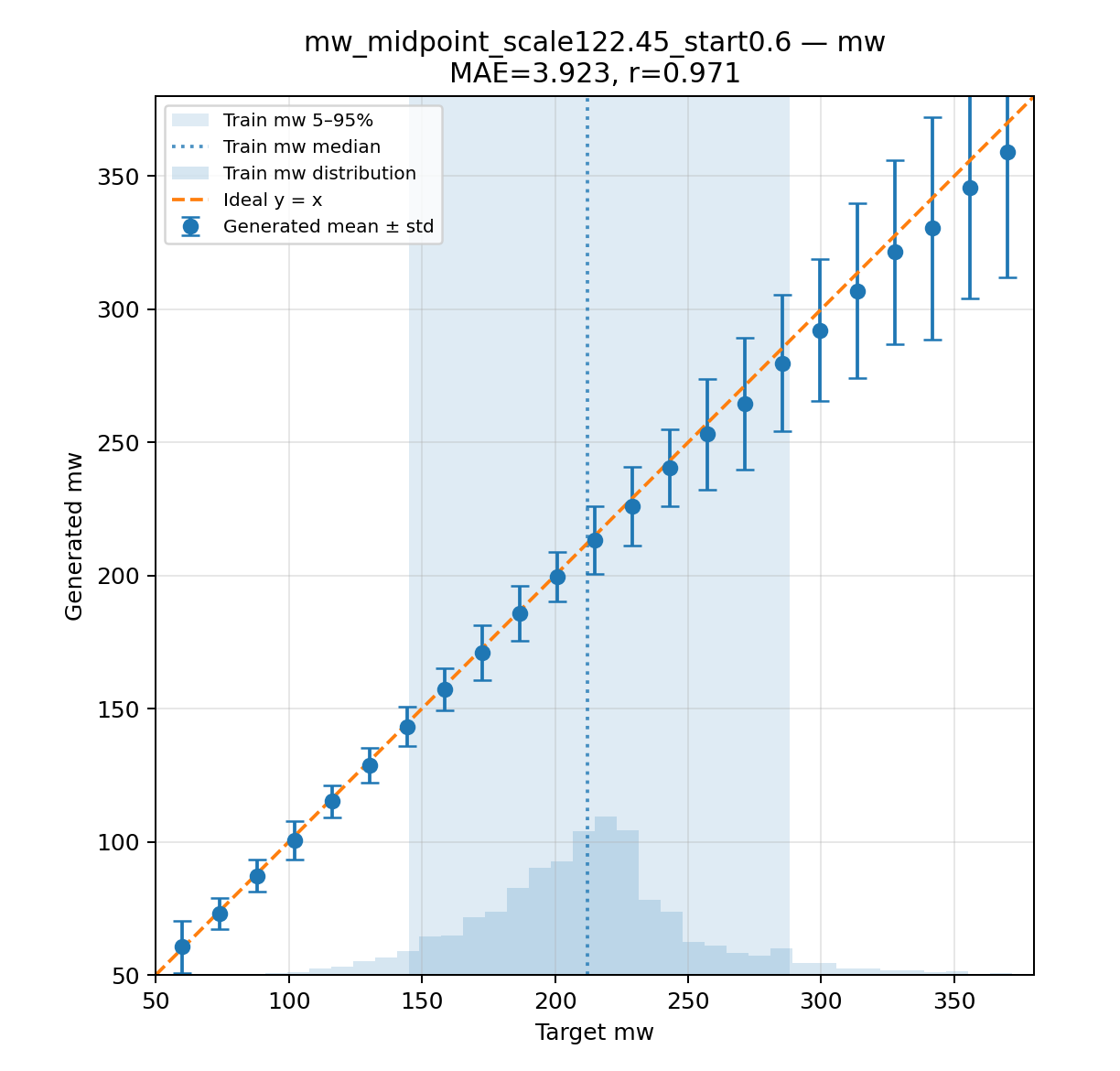}
        \caption*{(b) Molecular weight}
    \end{minipage}

    \caption{
    Independent property-controlled generation in the deterministic GraViti AE latent space on PubChem16. Generated properties are shown as a function of
the requested target for logP and molecular weight. The dashed line denotes ideal control ($y=x$), and the shaded region indicates the corresponding
training-distribution range.
    }
    \label{fig:ae_property_guidance}
\end{figure}

Figure~\ref{fig:ae_property_guidance} reports the target-response curves obtained in the deterministic AE space. The generated properties closely follow the requested targets for both logP and molecular weight, showing that the representation originally learned for unconditional graph generation already provides a suitable continuous domain for property steering. This supports the main premise of our framework: controlled generation does not require learning a separate property-specific encoder-decoder.

The variational representation nevertheless improves target fidelity, as summarized in Table~\ref{tab:ae_vae_guidance} in the main text. The MAE decreases from $0.247$ to $0.154$ for logP and from $3.923$ to $2.578$ for MW, together with increased Pearson correlations. These results suggest that the VAE latent space is more amenable to continuous property steering. This observation is also consistent with the molecular property-optimization experiments reported for GraViti, which found the variational latent representation to provide a favorable domain for navigating molecular properties.

\clearpage
\section{Validity under Property Control}
\label{app:local_validity}

\begin{figure}[ht]
    \centering


    \begin{subfigure}[t]{0.48\textwidth}
        \centering
        \includegraphics[width=\linewidth]{
        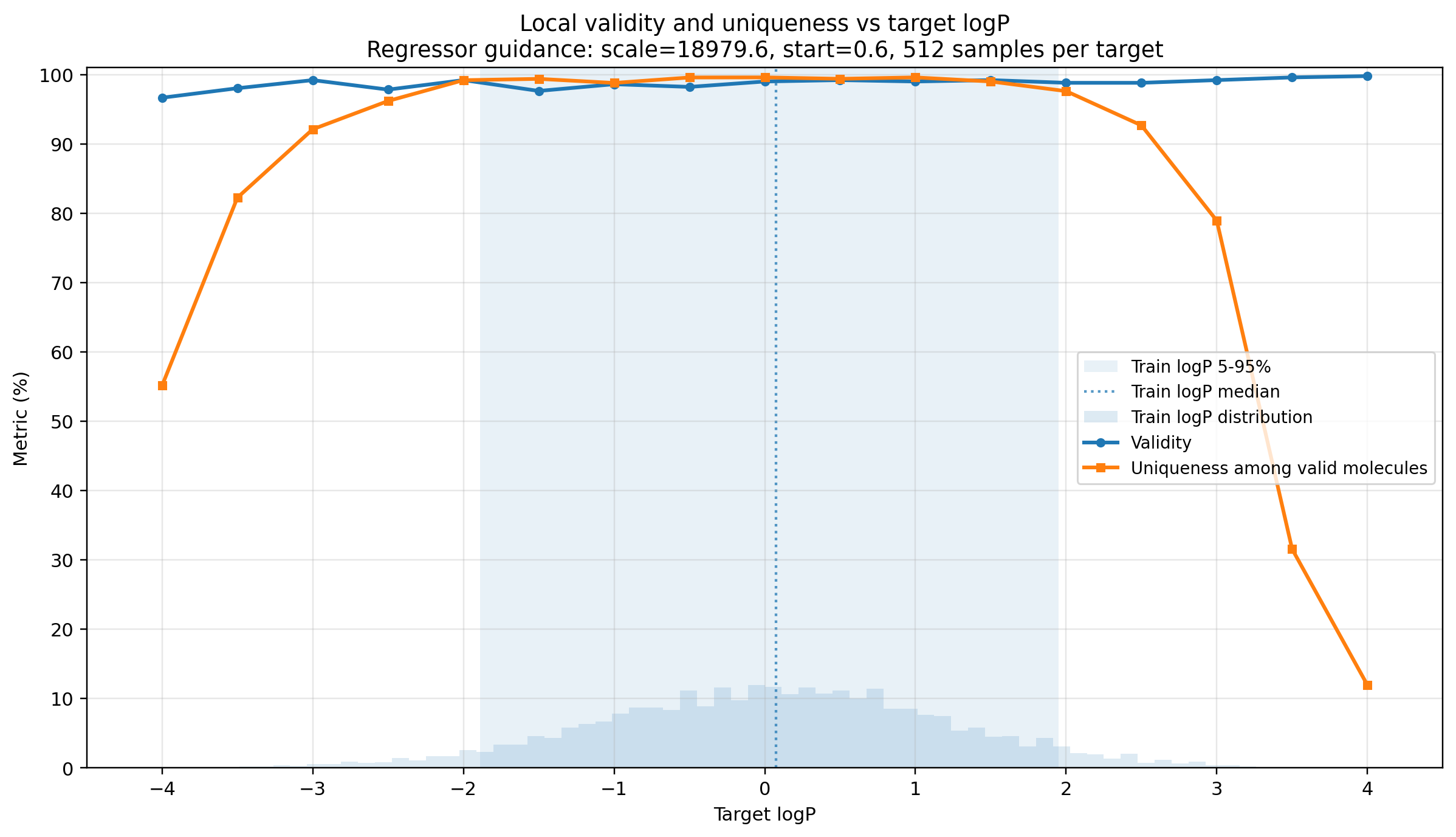
        }
        \caption{QM9 -- logP}
    \end{subfigure}
    \hfill
    \begin{subfigure}[t]{0.48\textwidth}
        \centering
        \includegraphics[width=\linewidth]{
        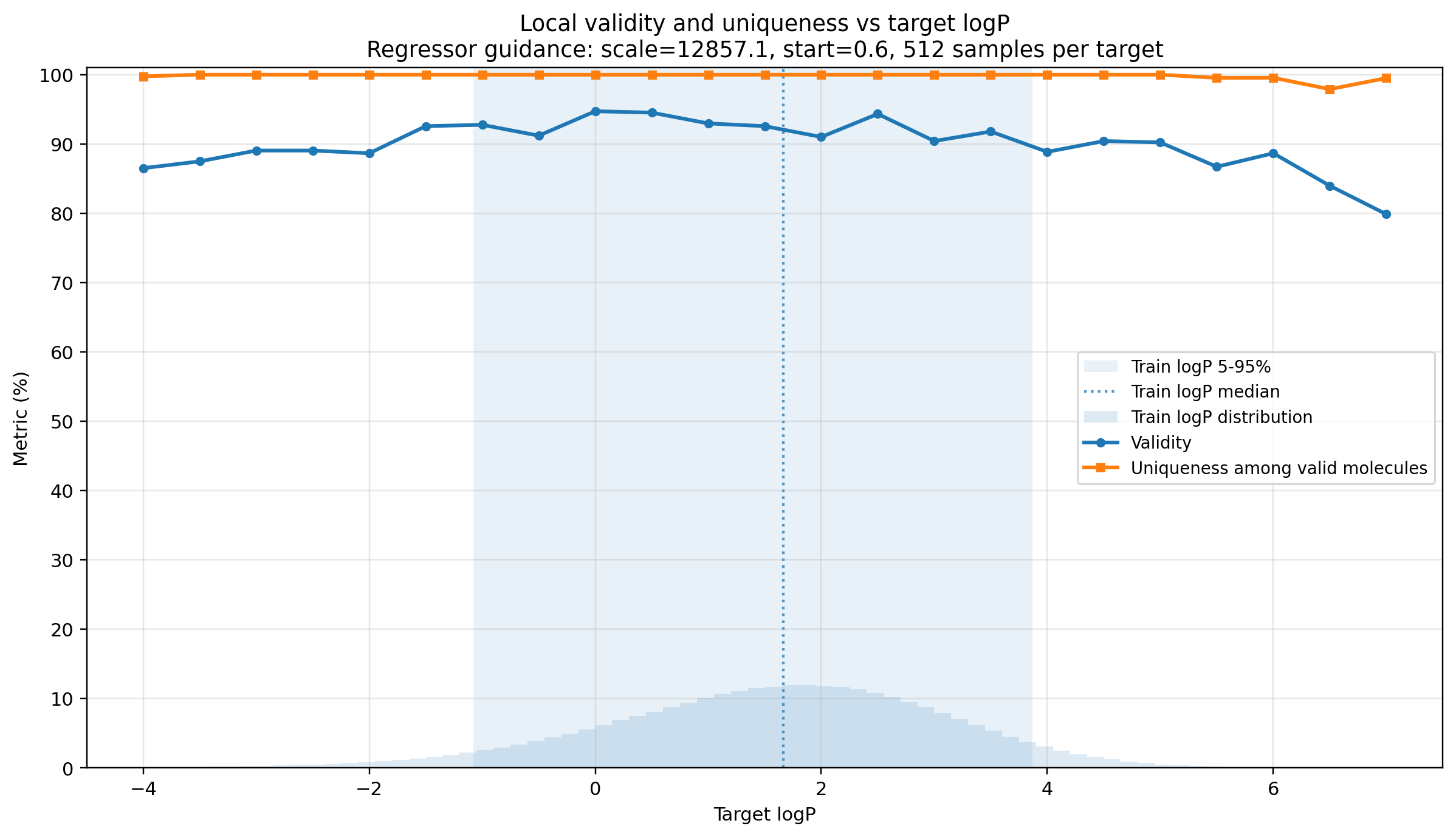
        }
        \caption{PubChem16 -- logP}
    \end{subfigure}

    \vspace{0.4cm}


    \begin{subfigure}[t]{0.48\textwidth}
        \centering
        \includegraphics[width=\linewidth]{
        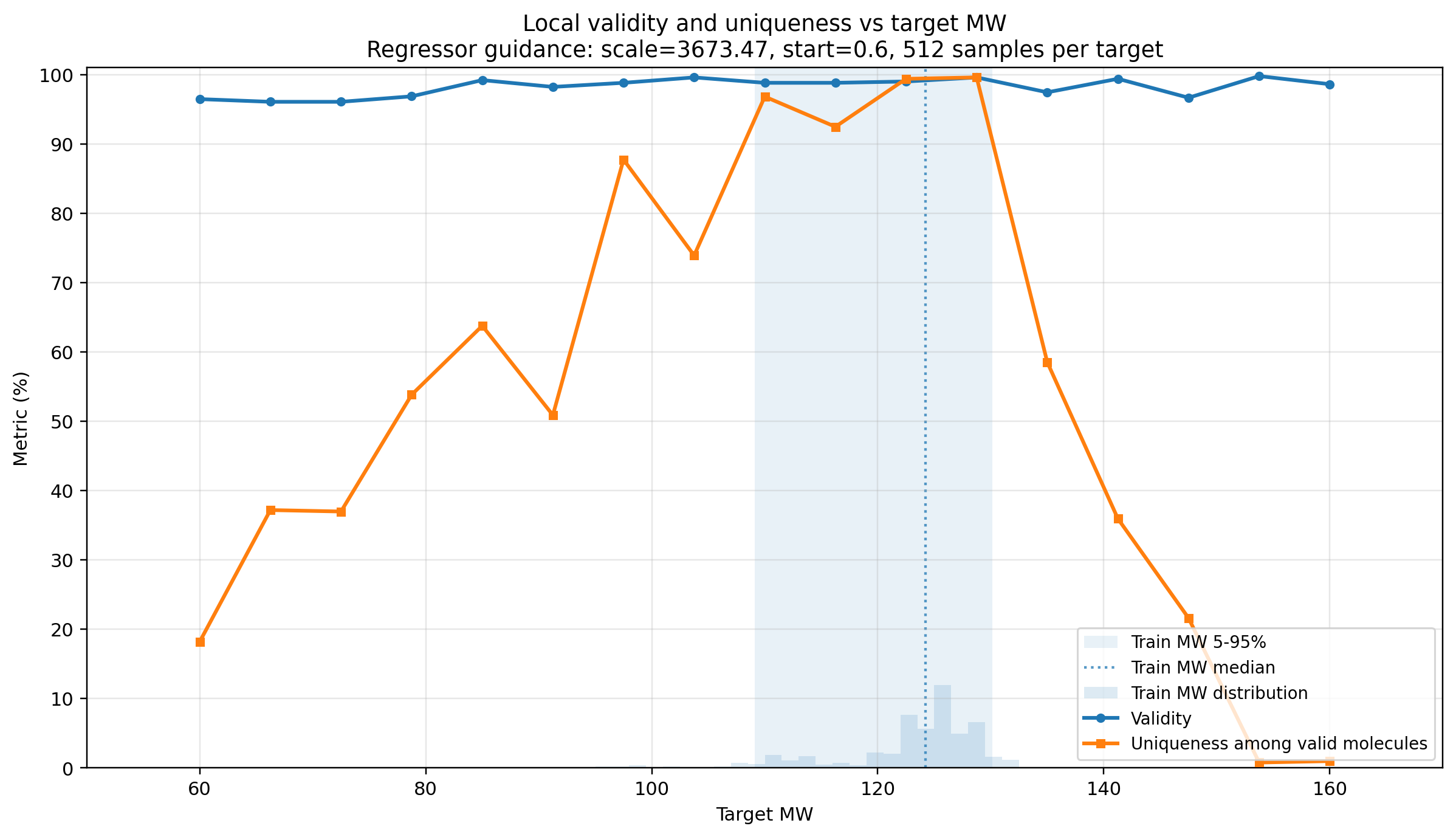
        }
        \caption{QM9 -- Molecular Weight}
    \end{subfigure}
    \hfill
    \begin{subfigure}[t]{0.48\textwidth}
        \centering
        \includegraphics[width=\linewidth]{
        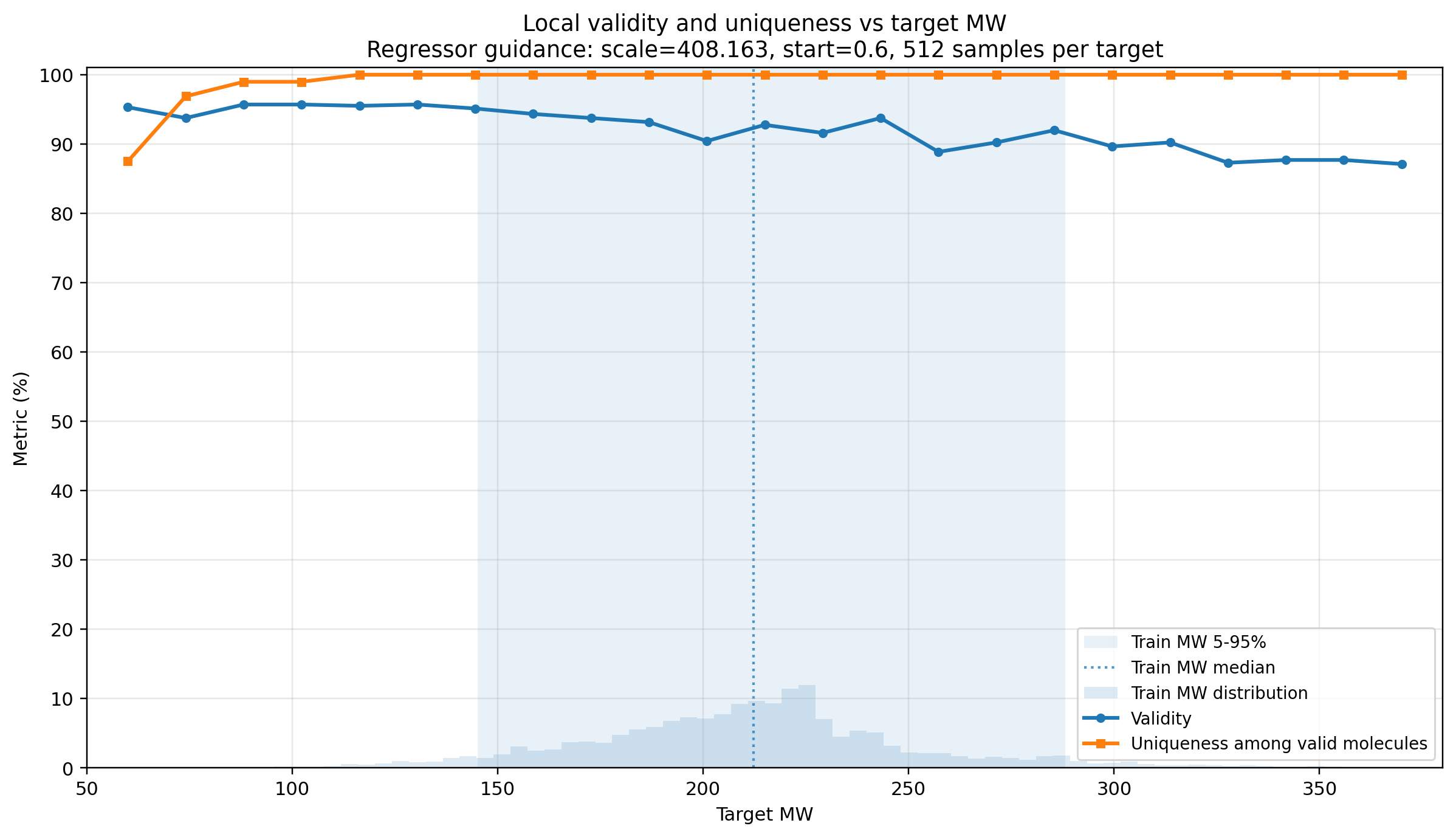
        }
        \caption{PubChem16 -- Molecular Weight}
    \end{subfigure}

    \caption{
    Validity and uniqueness as a function of the target property for
    property-guided molecular generation on QM9 and PubChem16.
    }
    \label{fig:validity_uniqueness_property_guidance}
\end{figure}

Figure~\ref{fig:validity_uniqueness_property_guidance} shows how validity and
uniqueness evolve with the requested property. Both remain high within the
training distribution, confirming that property control does not substantially
degrade generation quality over the range observed during training. Validity
decreases mainly for rarer or more extreme targets, revealing where stronger
steering begins to move samples toward latent regions that decode less reliably.
These results also qualitatively suggest that the models retain some robustness
beyond the training distribution.

\clearpage
\section{Extended CFM Training on PubChem16}
\label{app:pc16_cfm_30epochs}

To assess whether the performance of Categorical Flow Maps on PubChem16
is primarily limited by training duration, we additionally train the CSD and
ECLD variants for 30 epochs, corresponding to approximately 107 GPU hours.
Table~\ref{tab:pc16-cfm-30epochs} reports the resulting unconditional
generation performance across different numbers of sampling steps.

\begin{table}[h]
    \centering
    \caption{
    Unconditional generation results for CFM-CSD and CFM-ECLD on PubChem16
    after 30 training epochs ($\sim$107 GPU hours).
    Validity, uniqueness, and novelty are reported as mean $\pm$ standard
    deviation.
    }
    \label{tab:pc16-cfm-30epochs}
    \begin{tabular}{llcccc}
        \toprule
        \textbf{Method} &
        \textbf{Steps} &
        \textbf{Validity} $\uparrow$ &
        \textbf{Uniqueness} $\uparrow$ &
        \textbf{Novelty} $\uparrow$ &
        \textbf{FCD} $\downarrow$ \\
        \midrule

        CFM-CSD
        & 1
        & $88.19 \pm 1.11$
        & $99.51 \pm 0.23$
        & $75.10 \pm 1.30$
        & $6.8832$ \\

        CFM-CSD
        & 2
        & $86.48 \pm 0.97$
        & $99.98 \pm 0.07$
        & $87.03 \pm 1.01$
        & $2.5352$ \\

        CFM-CSD
        & 5
        & $93.85 \pm 0.75$
        & $100.00 \pm 0.00$
        & $88.58 \pm 0.93$
        & $1.4919$ \\

        CFM-CSD
        & 20
        & $97.58 \pm 0.51$
        & $99.99 \pm 0.02$
        & $88.77 \pm 1.00$
        & $0.9430$ \\

        \midrule

        CFM-ECLD
        & 1
        & $96.02 \pm 0.55$
        & $98.87 \pm 0.35$
        & $64.62 \pm 1.47$
        & $6.8654$ \\

        CFM-ECLD
        & 2
        & $94.45 \pm 0.64$
        & $99.96 \pm 0.06$
        & $85.91 \pm 1.27$
        & $2.6004$ \\

        CFM-ECLD
        & 5
        & $96.75 \pm 0.62$
        & $99.99 \pm 0.02$
        & $88.20 \pm 0.78$
        & $1.8569$ \\

        CFM-ECLD
        & 20
        & $97.63 \pm 0.40$
        & $99.98 \pm 0.04$
        & $88.06 \pm 0.93$
        & $1.6251$ \\

        \bottomrule
    \end{tabular}
\end{table}

\clearpage
\section{Global Validity under Property Control}
\label{app:global_val}

Global validity evaluates property-guided generation when target properties are
sampled according to their empirical training distribution. Under this setting,
property guidance preserves high validity and diversity on both datasets, as
reported in Table~\ref{tab:property_global_validity}. On QM9, validity remains
close to the unconditional regime for both properties, while uniqueness also
remains high. On PubChem16, validity stays above $0.92$ for both logP and MW
and uniqueness remains maximal.
\begin{table}[!h]
\centering
\caption{
Global validity and uniqueness under property-guided generation.
Target properties are sampled according to their empirical training
distribution before generating the corresponding molecules.
}
\label{tab:property_global_validity}
\begin{tabular}{llcc}
\toprule
\textbf{Dataset} &
\textbf{Property} &
\textbf{Validity} $\uparrow$ &
\textbf{Uniqueness} $\uparrow$ \\
\midrule

QM9
& logP
& $0.9873$
& $0.996$ \\

QM9
& MW
& $0.9824$
& $0.998$ \\

\midrule

PubChem16
& logP
& $0.9277$
& $1.000$ \\

PubChem16
& MW
& $0.9268$
& $1.000$ \\

\bottomrule
\end{tabular}
\end{table}

\clearpage
\section{Flow Matching}\label{app:flowmatching}

\subsection{Basic principle}

Flow Matching~\citep{lipman2022flow} learns a continuous transport between a
simple source distribution $p_0$ and a target distribution $p_1$. In our
setting, $p_1$ is the distribution of encoded graph representations. Given
$z_0 \sim p_0$, $z_1 \sim p_1$, and $t \sim \mathcal{U}[0,1]$, we consider the
linear probability path
\begin{equation}
    z_t = (1-t)z_0 + tz_1,
\end{equation}
with conditional velocity
\begin{equation}
    u_t(z_t \mid z_0,z_1) = z_1 - z_0.
\end{equation}
A neural vector field $v_{\theta}(z_t,t)$ is trained through the conditional
flow matching objective
\begin{equation}
    \mathcal{L}_{\mathrm{FM}}(\theta)
    =
    \mathbb{E}_{t,z_0,z_1}
    \left[
        \left\|
        v_{\theta}(z_t,t) - (z_1-z_0)
        \right\|_2^2
    \right].
\end{equation}
At inference time, samples are generated by integrating the learned vector
field from $t=0$ to $t=1$ and decoding the resulting latent representations.
We additionally consider mini-batch optimal-transport couplings between source
and target samples, following previous flow matching work, to construct more
coherent training paths.

\subsection{Guided generation}

A continuous latent representation also provides a direct interface for
controlled generation. Conditional flow matching augments the vector field with
a target property $c$,
\begin{equation}
    v_{\theta}(z_t,t,c),
\end{equation}
while post-hoc guidance can further modify the sampling dynamics using gradients
from a differentiable predictor. Given a predictor $r_{\omega}(z_t,t)$ and a
target $c$, the guided field takes the form
\begin{equation}
    \tilde{v}_{\theta}(z_t,t,c)
    =
    v_{\theta}(z_t,t,c)
    -
    \lambda
    \nabla_{z_t}
    \mathcal{L}_{\mathrm{prop}}
    \left(r_{\omega}(z_t,t),c\right),
\end{equation}
where $\lambda$ controls the guidance strength. This follows the general
principle of gradient-based generative guidance
\citep{dhariwal2021diffusion} and enables graph-level properties to be
controlled directly through the continuous latent dynamics.

\subsection{ODE Solvers}
\label{app:ode_solvers}

All latent flow models are sampled by numerically integrating the learned ordinary differential equation
\begin{equation}
\frac{d z_t}{dt} = v(z_t,t),
\qquad t\in[0,1],
\end{equation}
where $v$ denotes the learned velocity field, possibly augmented with a guidance term. We consider three explicit integration schemes: Euler, midpoint, and Heun. Let $z_n$ denote the current latent state at time $t_n$ and let $h=t_{n+1}-t_n$.

\paragraph{Euler.}
The Euler method performs a single evaluation of the velocity field per integration step:
\begin{equation}
z_{n+1}
=
z_n + h\,v(z_n,t_n).
\end{equation}

\paragraph{Midpoint.}
The midpoint method is a second-order Runge--Kutta scheme. It first evaluates the velocity at the beginning of the step,
\begin{equation}
k_1 = v(z_n,t_n),
\end{equation}
and uses it to estimate the state at the midpoint,
\begin{equation}
z_{n+\frac12}
=
z_n + \frac{h}{2}k_1.
\end{equation}
The velocity is then evaluated at this intermediate state,
\begin{equation}
k_2 =
v\!\left(z_{n+\frac12},t_n+\frac{h}{2}\right),
\end{equation}
and the update is given by
\begin{equation}
z_{n+1}=z_n+h\,k_2.
\end{equation}

\paragraph{Heun.}
Heun's method is also a second-order Runge--Kutta scheme. It first computes an Euler prediction,
\begin{align}
k_1 &= v(z_n,t_n),\
\tilde  z_{n+1} = z_n+h\,k_1,
\end{align}
then evaluates the velocity at the predicted endpoint,
\begin{equation}
k_2=v(\tilde z_{n+1},t_{n+1}),
\end{equation}
and averages the two slopes:
\begin{equation}
z_{n+1}
=
z_n+\frac{h}{2}(k_1+k_2).
\end{equation}

\paragraph{Number of function evaluations.}
Euler requires one velocity-field evaluation per integration step, whereas midpoint and Heun require two. Consequently, $S$ Euler steps correspond to $S$ NFE, while $S$ midpoint or Heun steps correspond to $2S$ NFE.

\clearpage 

\section{Qualitative examples on PubChem32}
\label{app:pc32_qualitative}

\begin{figure}[h]
    \centering

    \begin{subfigure}[t]{0.95\textwidth}
        \centering
        \includegraphics[width=\linewidth]{
        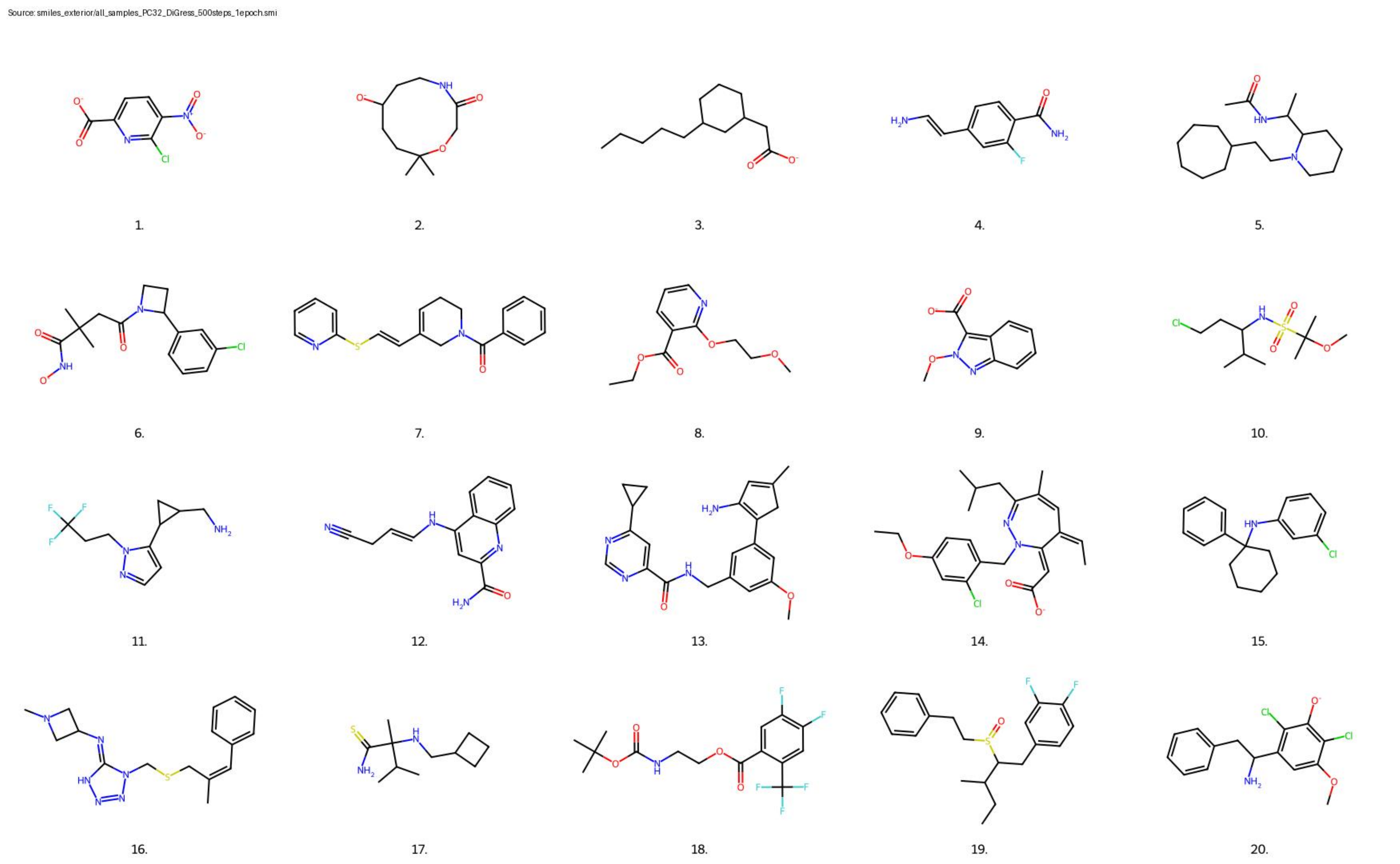
        }
        \caption{20 molecules generated by DiGress with 500 sampling steps.}
        \label{fig:pc32_digress_molecules}
    \end{subfigure}

    \vspace{0.5cm}

    \begin{subfigure}[t]{0.95\textwidth}
        \centering
        \includegraphics[width=\linewidth]{
        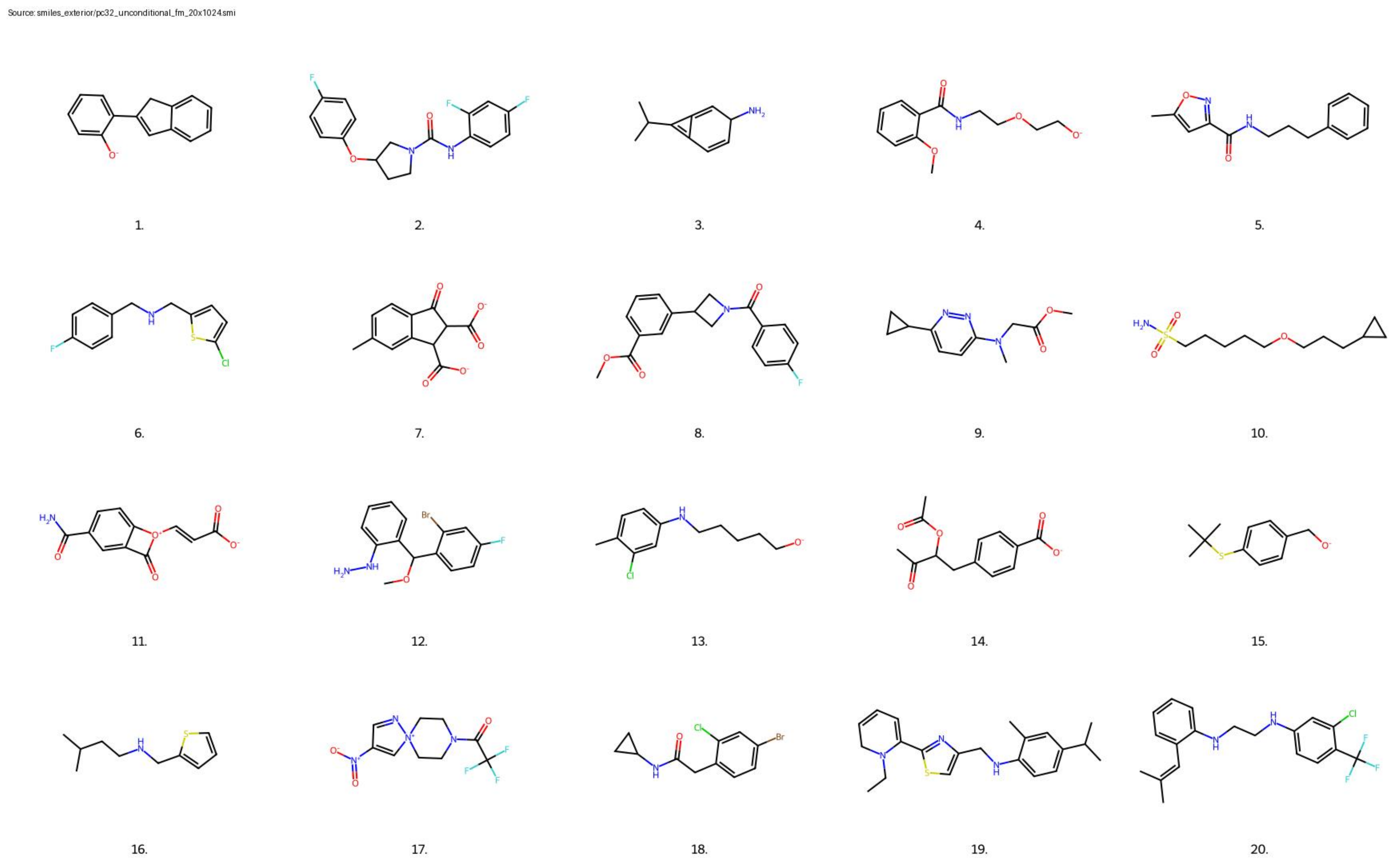
        }
        \caption{20 molecules generated by our Flow Matching model with 20 sampling steps.}
        \label{fig:pc32_fm_molecules}
    \end{subfigure}

    \caption{
    Qualitative examples of molecules generated on PubChem32 by
    DiGress (top) and our Flow Matching model (bottom).
    }
    \label{fig:pc32_generated_molecules}
\end{figure}

Figure~\ref{fig:pc32_generated_molecules} shows 20 example molecules
generated on PubChem32 using DiGress and our latent Flow Matching model.

\clearpage
\section{Examples of Property-Guided Generation}
\label{app:property_guided_examples}

\begin{sidewaysfigure}[h]
    \centering
    \includegraphics[width=1\textwidth]{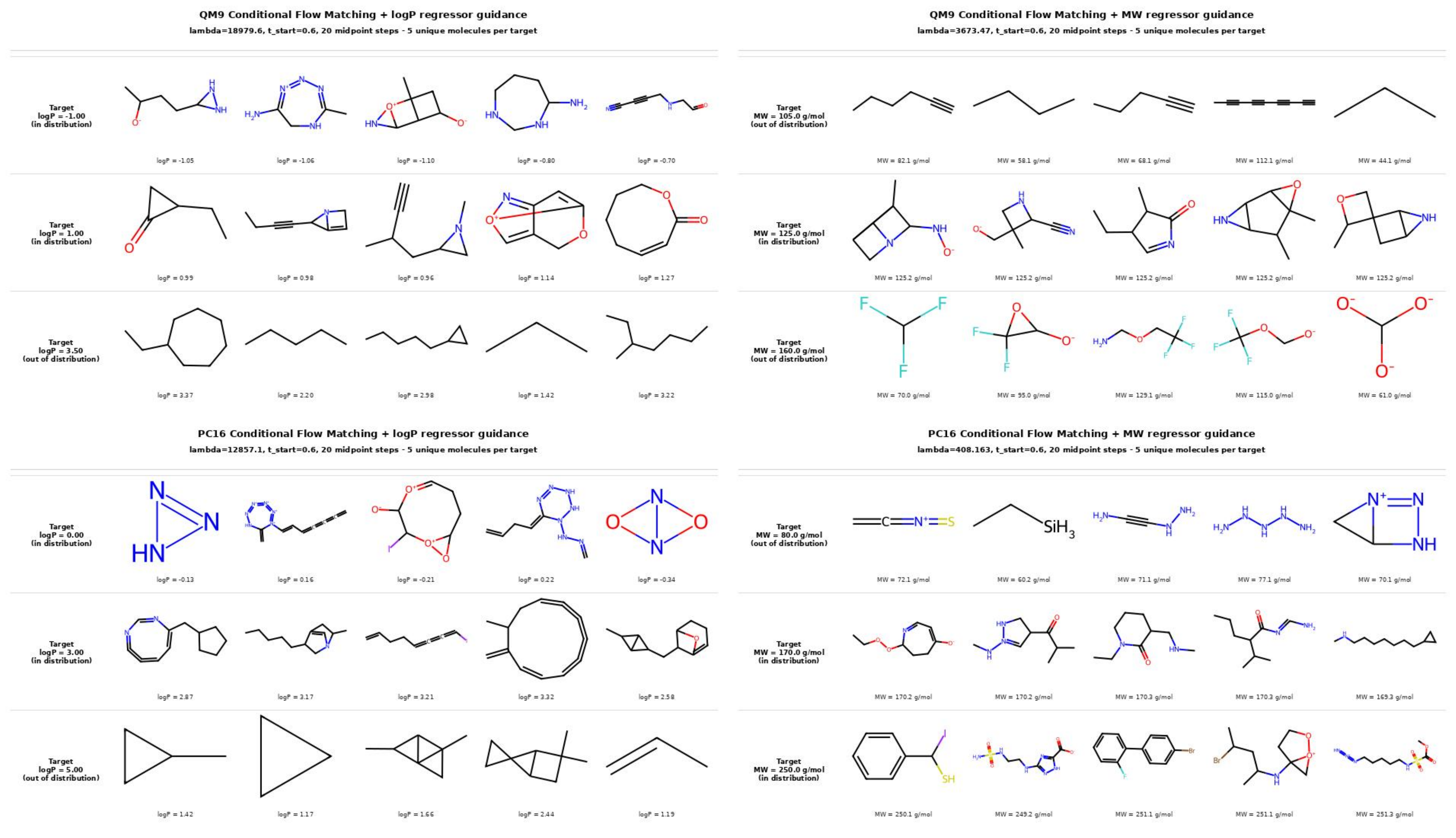}
    \caption{
    Examples of molecules generated with property-guided latent Flow Matching
    on QM9 and PubChem16 for logP and molecular weight (MW) targets.
    Examples are shown for both in-distribution and out-of-distribution target
    values. The achieved property value is reported below each generated
    molecule.
    }
    \label{fig:property_guided_examples}
\end{sidewaysfigure}

Figure~\ref{fig:property_guided_examples} provides qualitative examples of
molecules generated under property guidance for molecular weight (MW) and
logP. For targets within the training distribution, generated molecules
generally remain close to the requested property value. As targets move
outside the training distribution, larger deviations from the target can be
observed, illustrating the increased difficulty of extrapolating property
control beyond the training regime.

\clearpage

\section{Influence of the Latent Representation on Property Control}
\label{app:latent_representation_guidance}

We further investigate whether the choice of latent representation affects
property controllability. On PubChem16, we apply the same property-guidance
protocol to the deterministic and variational GraViti latent spaces. This
comparison isolates the effect of the representation itself, as both spaces
are reused directly from the unconditional generation setting, without
training a property-specific encoder or decoder.

As shown in Table~\ref{tab:ae_vae_guidance}, both representations support
effective control of logP and molecular weight (MW). However, the variational
representation consistently provides more accurate target matching, achieving
lower MAE and higher Pearson correlation for both properties.

\begin{table}[h]
\centering
\caption{
Effect of the graph-level latent representation on independent property control
on PubChem16. Both representations are evaluated using the same guidance
protocol.
}
\label{tab:ae_vae_guidance}
\begin{tabular}{llcc}
\toprule
\textbf{Representation} &
\textbf{Property} &
\textbf{MAE} $\downarrow$ &
$\mathbf{r}$ $\uparrow$ \\
\midrule

AE
& logP
& 0.247
& 0.966 \\

VAE
& logP
& \textbf{0.154}
& \textbf{0.978} \\

\midrule

AE
& MW
& 3.923
& 0.971 \\

VAE
& MW
& \textbf{2.578}
& \textbf{0.977} \\

\bottomrule
\end{tabular}
\end{table}

These results suggest that the variational representation is more amenable to
continuous property steering. This observation is consistent with the
property-optimization results reported for GraViti, where the variational
latent representation similarly provides a favorable space for navigating
molecular properties.

\clearpage
\section{Training settings}
\label{app:training_settings}

Table~\ref{tab:training_settings} summarizes the main architecture, training,
and sampling hyperparameters used for the models retained in our experiments.

\subsection{Vector Field Network}

We parameterize the velocity field $v_\theta(x, t)$ with a residual multilayer perceptron operating directly on the VAE latent space, $x \in \mathbb{R}^{d}$. The scalar time $t \in [0,1]$ is first mapped to a sinusoidal embedding $\gamma(t) \in \mathbb{R}^{d_t}$ and refined by a two-layer MLP with SiLU activation, yielding a time representation $e_t$. The network is a stack of $L$ pre-norm residual blocks of hidden width $h$, each of the form
\begin{equation}
    \mathrm{Block}(z) = z + W_2\,\mathrm{Dropout}\big(\mathrm{SiLU}(W_1\,\mathrm{LN}(z))\big),
\end{equation}
where $\mathrm{LN}$ denotes layer normalization. Time conditioning is injected in one of two ways. In the \emph{plain} variant, $e_t$ is concatenated to $x$ before a linear input projection to dimension $h$, so time enters only at the input. In the \emph{FiLM} variant, $x$ is projected on its own, and each block conditions on time through feature-wise linear modulation \cite{perez2018film}:
\begin{equation}
    (s_\ell, b_\ell) = W^{\mathrm{FiLM}}_\ell\, e_t, \qquad
    \tilde z = \mathrm{LN}(z) \odot (1 + s_\ell) + b_\ell,
\end{equation}
followed by the same two-layer feed-forward path and skip connection, so that time information is available at every depth. A final layer normalization and linear projection map the hidden state back to $\mathbb{R}^{d}$, producing a velocity of the same dimension as the latent.

The model is trained with the conditional flow matching objective on normalized latents. Given a noise sample $x_0 \sim \mathcal{N}(0, I)$, a data latent $x_1$, and $t \sim \mathcal{U}[0,1]$, we form the linear interpolation $x_t = (1-t)\,x_0 + t\,x_1$ and minimize
\begin{equation}
    \mathcal{L}(\theta) = \mathbb{E}_{t,\,(x_0, x_1)}\Big[\big\| v_\theta(x_t, t) - (x_1 - x_0) \big\|_2^2\Big].
\end{equation}
Optionally, the pairs $(x_0, x_1)$ are drawn from a Sinkhorn-regularized minibatch optimal transport plan rather than independently, which yields straighter trajectories. At inference, samples are generated by integrating $\dot{x} = v_\theta(x, t)$ from $t=0$ (Gaussian noise) to $t=1$ with an ODE solver (e.g., Heun's method), then denormalizing and decoding with the VAE decoder.

\begin{table}[h]
\centering
\small
\setlength{\tabcolsep}{12pt}
\renewcommand{\arraystretch}{1.08}

\caption{Hyperparameters used for the models retained in our experiments.}
\label{tab:training_settings}

\begin{tabular}{lcc}
\toprule
\textbf{Hyperparameter} & \textbf{QM9NoHydro} & \textbf{PubChem16} \\
\midrule

Latent dimension
& 64
& 256 \\

Latent transform (FM)
& None
& Standardization \\

\midrule
\multicolumn{3}{l}{\textit{Flow Matching}} \\

Hidden dimension
& 1024
& 2048 \\

Number of layers
& 6
& 10 \\

Time embedding dimension
& 1
& 64 \\

FiLM
& Yes
& Yes \\

Batch OT
& Yes
& Yes \\

Dropout
& 0
& 0 \\

Optimizer
& AdamW
& AdamW \\

Learning rate
& $2\times10^{-4}$
& $2\times10^{-4}$ \\

Weight decay
& $10^{-5}$
& $10^{-5}$ \\

Batch size
& 1024
& 1024 \\

Training epochs
& 500
& 30 \\

Warmup steps
& 1000
& 1000 \\

Gradient clipping
& 1
& 1 \\

\midrule
\multicolumn{3}{l}{\textit{Validity predictor}} \\

Hidden dimension
& 256
& 1024 \\

Number of layers
& 3
& 6 \\

Time conditioning
& Concatenated to input
& Concatenated to input \\

Learning rate
& $3\times10^{-4}$
& $3\times10^{-4}$ \\

Weight decay
& $10^{-5}$
& $10^{-5}$ \\

Batch size
& 1024
& 1024 \\

Maximum epochs
& 300
& 300 \\

Selected epoch
& 266
& 50 \\

Early-stopping patience
& 50
& 30 \\

\midrule
\multicolumn{3}{l}{\textit{Validity Flow Matching}} \\

Latent transform
& Standardization
& Standardization \\

Hidden dimension
& 1536
& 2048 \\

Number of layers
& 8
& 10 \\

Dropout
& 0
& 0 \\

Log-variance bounds
& $[-6,2]$
& $[-6,2]$ \\

Learning rate
& $3\times10^{-4}$
& $3\times10^{-4}$ \\

Minimum learning rate
& $10^{-5}$
& $10^{-5}$ \\

Weight decay
& $10^{-5}$
& $10^{-5}$ \\

Batch size
& 512
& 1024 \\

Maximum epochs
& 500
& 150 \\

Selected epoch
& 500
& 32 \\

Early-stopping patience
& 500
& 5 \\

Endpoint candidates $K$
& 8
& 32 \\

Reweighting strength $\beta$
& 1
& 5 \\

\midrule
\multicolumn{3}{l}{\textit{Sampling}} \\

Integrator
& Euler
& Euler \\

Integration interval
& $[0,1]$
& $[0,1]$ \\

Default FM steps
& 500
& 20 \\

VFM steps
& 20 / 500
& 20 / 500 \\

Guidance scale
& 25
& 10 \\

Guidance start
& $t=0.6$
& $t=0.6$ \\

\bottomrule
\end{tabular}
\end{table}

\begin{table*}[h]
\centering
\small
\setlength{\tabcolsep}{8pt}
\renewcommand{\arraystretch}{1.10}
\caption{
Hyperparameters used for property-guided generation on QM9NoHydro.
Conditioning properties are standardized using training-set statistics. No latent
transformation is applied.
}
\label{tab:property_guidance_qm9}
\begin{tabular}{lcc}
\toprule
\textbf{Hyperparameter} & \textbf{QM9 MW} & \textbf{QM9 logP} \\
\midrule

Latent dimension
& 64 & 64 \\

Conditioning property
& MW & logP \\

Property normalization
& Standardized & Standardized \\

Latent transform
& None & None \\

\midrule
\multicolumn{3}{l}{\textit{Conditional Flow Matching}} \\

Hidden dimension
& 512 & 512 \\

Number of residual blocks
& 6 & 6 \\

Conditioning mechanism
& FiLM & FiLM \\

Condition dimension
& 1 & 1 \\

Dropout
& 0 & 0 \\

Batch OT
& Yes & Yes \\

Optimizer
& AdamW & AdamW \\

Learning rate
& $3\times10^{-4}$ & $3\times10^{-4}$ \\

Weight decay
& $10^{-5}$ & $10^{-5}$ \\

Batch size
& 512 & 512 \\

Training epochs
& 500 & 500 \\

Validation split
& 0.1 & 0.1 \\

Gradient clipping
& 0.5 & 0.5 \\

\midrule
\multicolumn{3}{l}{\textit{Property regressor}} \\

Hidden dimension
& 256 & 256 \\

Number of hidden layers
& 3 & 3 \\

Activation
& SiLU & SiLU \\

Normalization
& LayerNorm & LayerNorm \\

Time conditioning
& Concatenated to input & Concatenated to input \\

Output dimension
& 1 & 1 \\

Optimizer
& Adam & Adam \\

Learning rate
& $3\times10^{-4}$ & $3\times10^{-4}$ \\

Weight decay
& $10^{-5}$ & $10^{-5}$ \\

Training epochs
& 300 & 300 \\

Gradient clipping
& 1.0 & 1.0 \\

\midrule
\multicolumn{3}{l}{\textit{Batch optimal transport}} \\

Sinkhorn batch size
& 512 & 512 \\

Regularization
& 1.0 & 1.0 \\

Maximum iterations
& 1000 & 1000 \\

Convergence threshold
& $10^{-6}$ & $10^{-6}$ \\

\midrule
\multicolumn{3}{l}{\textit{Sampling and guidance}} \\

ODE solver
& Midpoint & Midpoint \\

Number of steps
& 20 & 20 \\

Temperature
& 1.0 & 1.0 \\

Guidance start
& $t=0.6$ & $t=0.6$ \\

Guidance scale
& 3673.47 & 18979.6 \\

\midrule

Random seed
& 42 & 42 \\

Samples per target
& 512 & 512 \\

Global evaluation samples
& 1024 & 1024 \\

\bottomrule
\end{tabular}
\end{table*}

\begin{table*}[h]
\centering
\small
\setlength{\tabcolsep}{8pt}
\renewcommand{\arraystretch}{1.10}
\caption{
Hyperparameters used for property-guided generation on PubChem16.
Conditioning properties are standardized using training-set statistics. No latent
transformation is applied.
}
\label{tab:property_guidance_pubchem16}
\begin{tabular}{lcc}
\toprule
\textbf{Hyperparameter} & \textbf{PubChem16 MW} & \textbf{PubChem16 logP} \\
\midrule

Latent dimension
& 256 & 256 \\

Conditioning property
& MW & logP \\

Property normalization
& Standardized & Standardized \\

Latent transform
& None & None \\

\midrule
\multicolumn{3}{l}{\textit{Conditional Flow Matching}} \\

Hidden dimension
& 1024 & 1024 \\

Number of residual blocks
& 6 & 6 \\

Conditioning mechanism
& FiLM & FiLM \\

Condition dimension
& 1 & 1 \\

Dropout
& 0 & 0 \\

Batch OT
& Yes & Yes \\

Optimizer
& AdamW & AdamW \\

Learning rate
& $3\times10^{-4}$ & $3\times10^{-4}$ \\

Weight decay
& $10^{-5}$ & $10^{-5}$ \\

Batch size
& 512 & 512 \\

Training epochs
& 1000 & 1000 \\

Validation split
& 0.1 & 0.1 \\

Gradient clipping
& 0.5 & 0.5 \\

\midrule
\multicolumn{3}{l}{\textit{Property regressor}} \\

Hidden dimension
& 256 & 256 \\

Number of hidden layers
& 3 & 3 \\

Activation
& SiLU & SiLU \\

Normalization
& LayerNorm & LayerNorm \\

Time conditioning
& Concatenated to input & Concatenated to input \\

Output dimension
& 1 & 1 \\

Optimizer
& Adam & Adam \\

Learning rate
& $3\times10^{-4}$ & $3\times10^{-4}$ \\

Weight decay
& $10^{-5}$ & $10^{-5}$ \\

Training epochs
& 300 & 300 \\

Gradient clipping
& 1.0 & 1.0 \\

\midrule
\multicolumn{3}{l}{\textit{Batch optimal transport}} \\

Sinkhorn batch size
& 512 & 512 \\

Regularization
& 1.0 & 1.0 \\

Maximum iterations
& 1000 & 1000 \\

Convergence threshold
& $10^{-6}$ & $10^{-6}$ \\

\midrule
\multicolumn{3}{l}{\textit{Sampling and guidance}} \\

ODE solver
& Midpoint & Midpoint \\

Number of steps
& 20 & 20 \\

Temperature
& 1.0 & 1.0 \\

Guidance start
& $t=0.6$ & $t=0.6$ \\

Guidance scale
& 408.163 & 12857.1 \\

\midrule

Random seed
& 42 & 42 \\

Samples per target
& 512 & 512 \\

Global evaluation samples
& 1024 & 1024 \\

\bottomrule
\end{tabular}
\end{table*}
\end{document}